\documentclass[graybox]{svmult}

\usepackage{type1cm}        

\usepackage{makeidx}         
\usepackage{graphicx}        
\graphicspath{{figures/}}
\DeclareGraphicsExtensions{.pdf,.jpeg,.png}

\usepackage{multicol}        
\usepackage[bottom]{footmisc}

\usepackage{newtxtext}       %
\usepackage{newtxmath}       

\makeindex             

\usepackage{multirow}
\usepackage{rotating}
\usepackage[numbers,sort&compress]{natbib}
\usepackage{url}
\usepackage{xcolor}
\usepackage{booktabs}

\begin{document}

\title*{How to train accurate BNNs for embedded systems?}
\author{Floran de Putter and Henk Corporaal}
\institute{Floran de Putter \at Eindhoven Artificial Intelligence Systems Institute and PARsE lab, Eindhoven University of Technology, PO Box 513, Eindhoven 5600 MB, the Netherlands, \email{f.a.m.d.putter@tue.nl}
\and Henk Corporaal \at Eindhoven Artificial Intelligence Systems Institute and PARsE lab, Eindhoven University of Technology, PO Box 513, Eindhoven 5600 MB, the Netherlands, \email{h.corporaal@tue.nl}}
\maketitle

\abstract*{A key enabler of deploying convolutional neural networks on resource-constrained embedded systems is the binary neural network (BNN). BNNs save on memory and simplify computation by binarizing both features and weights. Unfortunately, binarization is inevitably accompanied by a severe decrease in accuracy. To reduce the accuracy gap between binary and full-precision networks, many repair methods have been proposed in the recent past, which we have classified and put into a single overview in this chapter. The repair methods are divided into two main branches, training techniques and network topology changes, which can further be split into smaller categories. The latter category introduces additional cost (energy consumption or additional area) for an embedded system, while the former does not. From our overview, we observe that progress has been made in reducing the accuracy gap, but BNN papers are not aligned on what repair methods should be used to get highly accurate BNNs. Therefore, this chapter contains an empirical review that evaluates the benefits of many repair methods in isolation over the ResNet-20\&CIFAR10 and ResNet-18\&CIFAR100 benchmarks. We found three repair categories most beneficial: feature binarizer, feature normalization, and double residual. Based on this review we discuss future directions and research opportunities. We sketch the benefit and costs associated with BNNs on embedded systems because it remains to be seen whether BNNs will be able to close the accuracy gap while staying highly energy-efficient on resource-constrained embedded systems.}
\abstract{A key enabler of deploying convolutional neural networks on resource-constrained embedded systems is the binary neural network (BNN). BNNs save on memory and simplify computation by binarizing both features and weights. Unfortunately, binarization is inevitably accompanied by a severe decrease in accuracy. To reduce the accuracy gap between binary and full-precision networks, many repair methods have been proposed in the recent past, which we have classified and put into a single overview in this chapter. The repair methods are divided into two main branches, training techniques and network topology changes, which can further be split into smaller categories. The latter category introduces additional cost (energy consumption or additional area) for an embedded system, while the former does not. From our overview, we observe that progress has been made in reducing the accuracy gap, but BNN papers are not aligned on what repair methods should be used to get highly accurate BNNs. Therefore, this chapter contains an empirical review that evaluates the benefits of many repair methods in isolation over the ResNet-20\&CIFAR10 and ResNet-18\&CIFAR100 benchmarks. We found three repair categories most beneficial: feature binarizer, feature normalization, and double residual. Based on this review we discuss future directions and research opportunities. We sketch the benefit and costs associated with BNNs on embedded systems because it remains to be seen whether BNNs will be able to close the accuracy gap while staying highly energy-efficient on resource-constrained embedded systems.}

\section{Introduction}
With the advent of deep learning, deep learning architectures have achieved state-of-the-art results in a variety of fields, e.g., computer vision, speech recognition, and natural language processing, in which they surpass (or are comparable to) human expert performance. Within the popular domain of computer vision, Convolutional Neural Networks (CNNs) have been proven successful and are thus widely used. Nevertheless, CNNs require massive amounts of memory and computational load for both training and inference. While training can be done offline on powerful GPU-based machines, inference is expected to be done at the edge on resource-constrained embedded systems, because of data privacy and strict latency and reliability constraints.

State-of-the-art CNNs have billions of parameters and do billions of operations, making them both computationally and memory intensive and therefore hard to deploy on resource-constrained embedded systems. Fortunately, there is a lot of room for optimization. Key optimizations are compression and code transformations. Compression reduces the model size of CNNs with a minor loss in  accuracy compared to the original model. High compression can be achieved by quantization and/or pruning. Quantization reduces the bit width for both parameters and features of a CNN, whereas pruning removes redundant connections and/or neurons in a CNN. Code transformations enable more energy-efficient mappings of CNNs on embedded systems. A proper CNN mapping makes sure many weights and features are cached on-chip and reused many times before new data is loaded from off-chip power-hungry DRAM. Therefore, a proper mapping is key to having CNNs on embedded systems.

The focus of this chapter is on extreme quantized CNNs, specifically: Binary Neural Networks (BNNs). This is a relatively new area since the first BNN \cite{Courbariaux2016Binarized-1} was published in 2016. BNNs achieve a very high rate of compression by quantizing both features and weights to a single bit. Thereby largely saving memory and simplifying computation enabling them to be the holy grail for deploying CNNs on resource-constrained embedded systems. However, this extreme form of quantization inescapably causes severe accuracy loss. Moreover, BNNs' intrinsic discontinuity brings difficulty to their training. Fortunately, many training techniques and network topology changes have been proposed that aim to reduce this accuracy loss. 

This chapter presents a survey of these training techniques and network topology changes that aim to repair the accuracy loss. Problem, however, is that papers on BNNs are not aligned on what improvements should be used to get high-accuracy BNNs. Typically, each paper on BNNs proposes its own improvements that are not necessarily in line, or even contradict, with previous work. Therefore, we isolate improvements and empirically evaluate those on benchmarks for image classification. Based on the outcomes of this study we provide directions for future research.

In short, the contributions of this work are:
\begin{itemize}
    \item Classification and overview of BNN accuracy repair methods under various dimensions.
    \item Evaluation of the benefits of isolated improvements on two benchmarks: ResNet-20\&CIFAR10 and ResNet-18\&CIFAR100.
    \item Indication of future directions for researching accuracy repair techniques that allow high-accuracy BNNs.
\end{itemize}
The remainder of this chapter is structured as follows: next Section reviews related work, while Section \ref{sec:background} presents background information on BNNs. Section \ref{sec:classification} and \ref{sec:overview} respectively present the classification and overview of accuracy repair methods. In Section \ref{sec:review} the empirical review is conducted and results are shown. Section \ref{sec:discussion} raises a discussion on the results and indicate future research directions. Finally, in Section \ref{sec:conclusion} conclusions are drawn.

\section{Related work}
This chapter consists of a survey and an empirical study. With respect to the survey part, there are three previously published surveys on BNNs \cite{Simons2019ANetworks,Qin2020BinarySurvey,Yuan2021ANetwork} in respectively \citeyear{Simons2019ANetworks}, \citeyear{Qin2020BinarySurvey} and \citeyear{Yuan2021ANetwork}. Since the BNN field is relatively new, it is quickly evolving and therefore many new papers are published each year. Therefore, we feel it is reasonable to present another overview that is updated with new accuracy repair methods. Contrary to the most recent survey \cite{Yuan2021ANetwork}, we employ a hierarchical classification with different metrics that allows us to present a complete overview in a single table in which each work may contain several accuracy repair techniques. In addition, none of the previously published surveys have done an empirical study on the benefits of individual repair methods.

With respect to the empirical study, there are four works \cite{Tang2017HowAccuracy,Alizadeh2019AnOptimisation,Bethge2019BinaryDenseNet:Networks,Liu2021HowOptimization} that research a limited set of our design space. \cite{Tang2017HowAccuracy} study the effect of certain learning rates and propose to use a regularization loss term for absolute latent weight distance to one instead of weight decay. Furthermore, they suggest using a PReLU activation function rather than a ReLU. \cite{Alizadeh2019AnOptimisation} studies the difference between the ADAM and SGD optimizer, the effect of gradient clipping, and the momentum term in batch normalization. \cite{Bethge2019BinaryDenseNet:Networks} conducts experiments on a scaling factor that is derived from features and weights. In addition, they study the clipping interval on the binarizer and use the double residual repair method in all experiments. \cite{Liu2021HowOptimization} aims to explain why the ADAM optimizer performs better than SGD in their experiments. Additionally, they investigate why the two-stage training scheme is beneficial in their experiments. Contrary to those studies, we first establish a good baseline without any repairs applied to it. Furthermore, each experiment in our design space is run five times with different random seeds, such that claims about the deviation in accuracy of a repair method can be made. Lastly, as our design space is a lot larger; our study aims to provide the full picture of the benefits of individual repair methods.

\section{Background on BNNs}\label{sec:background}
State-of-the-art CNNs need a lot of memory and compute power, which often makes them unsuitable for resource-constrained embedded systems. Quantization is one optimization that may enable CNNs on embedded systems. As the holy grail of quantization, BNNs achieve a 32x compression ratio, compared to 32-bit floating-point CNNs, by quantizing both features and weights to either $+1$ or $-1$ such that they can be stored in a single bit. Section \ref{sec:background:inference} explains the inference process, while Section \ref{sec:background:training} presents the training process of BNNs.

\subsection{Inference} \label{sec:background:inference}
To quantize features or weights from real to binary values generally the sign-function is used:
\begin{equation}\label{eqn:sign}
    x_B = \text{sign}(x) = \begin{cases}
    +1 & \text{if } x\ge 0\\
    -1 & \text{otherwise}
    \end{cases}
\end{equation}
where $x_B$ is the binarized value and $x$ the real value. 

\begin{figure}
  \centering
  \includegraphics[width=0.9\textwidth]{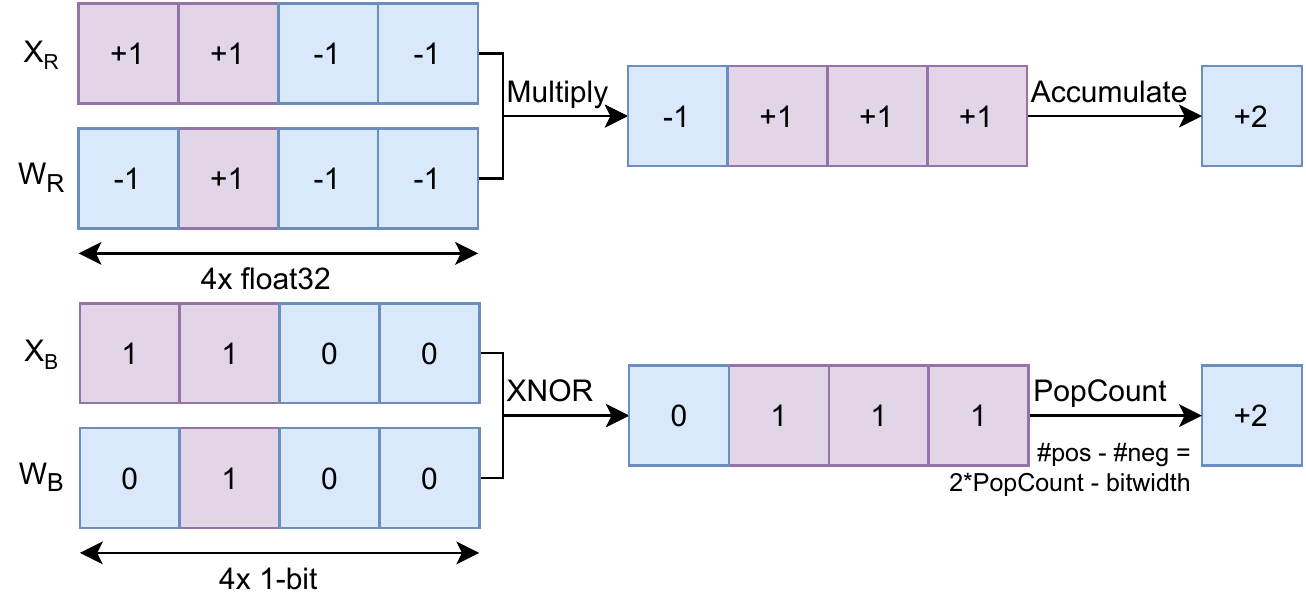}
  \caption{An example of the multiply-accumulate operation using binary values: a combination of bitwise XNOR and popcount.}
  \label{fig:mac_xnorpopcount}
\end{figure}
Not only does the binarization save on storage space, but it also simplifies the compute logic. Wherein a CNN's convolution layers (and others like fully-connected layers) are done using many real-valued multiplications and accumulations, a binary convolution can be implemented using XNORs and popcount operation (i.e. counting the number of $1$’s in a bitstring). An example of this is given in Fig. \ref{fig:mac_xnorpopcount}. Note that the $-1$ is encoded as a $0$. 

Replacing a real-valued MAC by XNOR-PopCount is very good for energy efficiency. E.g., in \cite{Andri2021ChewBaccaNN:Accelerator} a BNN accelerator is presented that achieves down to merely $4.48$ fJ/Op in GF 22nm at $0.4$V, where Op is a binary operation (xnor or popcount).

\begin{figure}
  \centering
  \includegraphics[width=\textwidth]{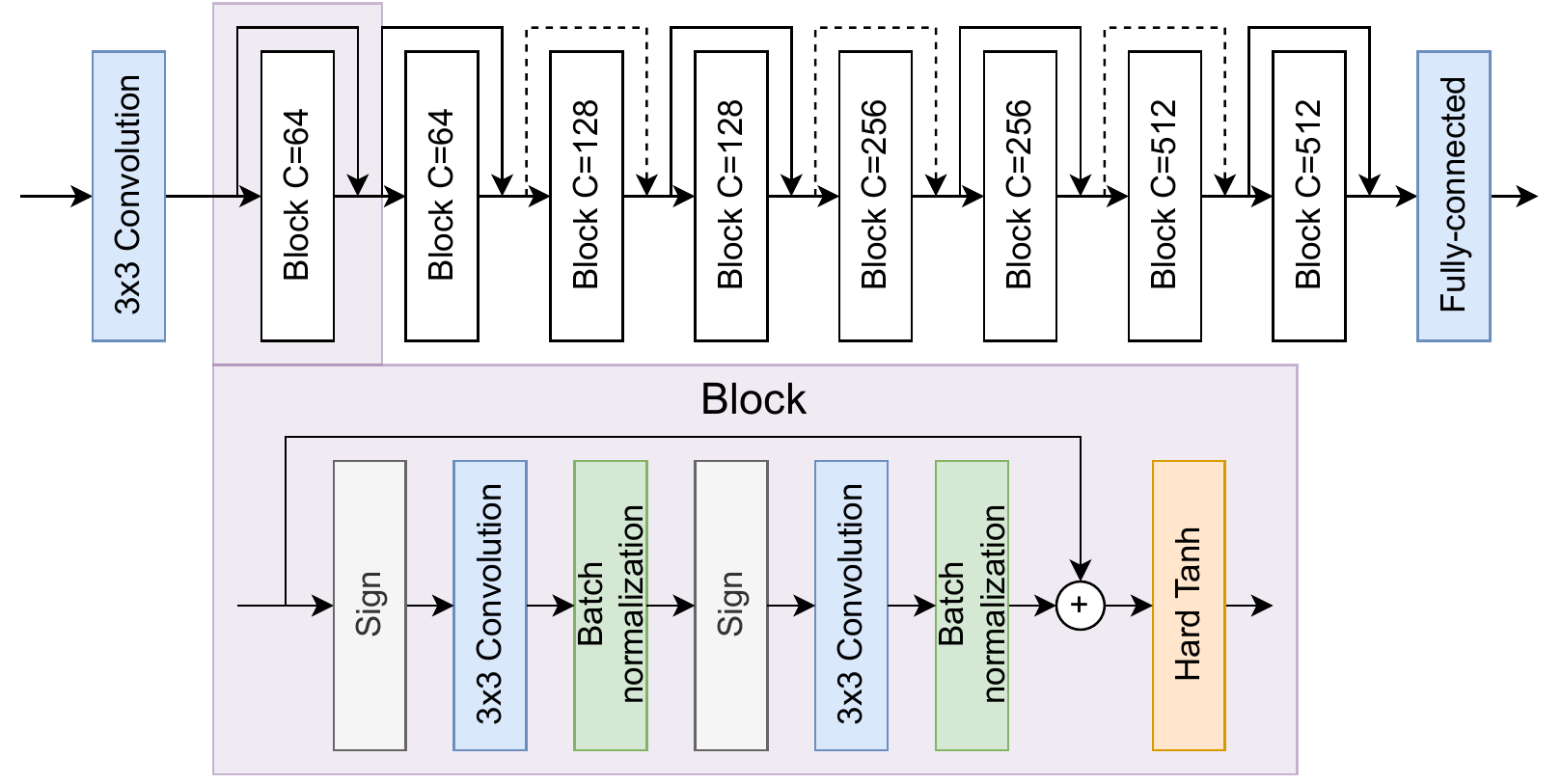}
  \caption{The binary ResNet-18 architecture is composed of multiple blocks. Each block consists of two convolutions and has a varying amount of input channels, denoted by C. Dashed residual lines indicate down-sampling of the height and width of the feature map. Moreover, the first convolution layer inside a block with the dashed residual line employs a stride of two.}
  \label{fig:resnet18}
\end{figure}

The most common network architecture used in BNN papers is illustrated in Fig. \ref{fig:resnet18}. It is based on the ResNet-18 \cite{He2015DeepRecognition} architecture which consists of a single building block that repeats. Note that each repetition has a different amount of input channels. This architecture is also used in our review in Section \ref{sec:review}.

\begin{figure}
  \centering
  \includegraphics[width=0.7\textwidth]{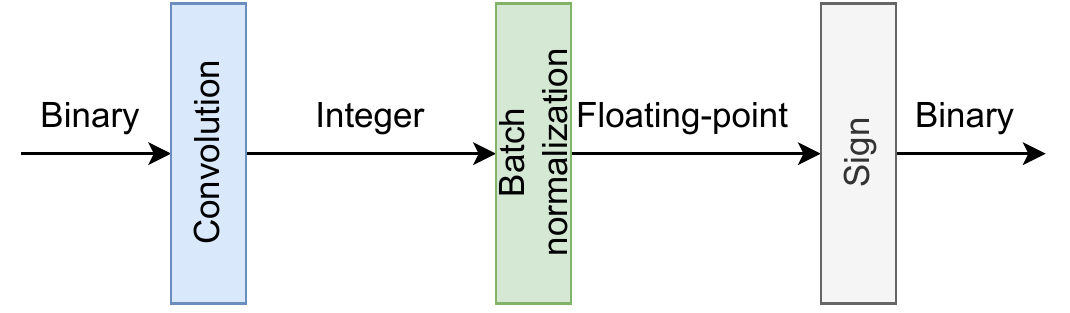}
  \caption{Common BNN layer sequence: convolution followed by batch normalization and sign-function. The text above the arrows describe the data format at that specific location in the network.}
  \label{fig:vgg}
\end{figure}

A common layer sequence of BNNs is shown in Fig. \ref{fig:vgg}. To optimize this part for inference BNNs typically fuse the batch normalization layer with a sign function, instead of the preceding convolution layer as done in full-precision networks. The absorption of the batch normalization layer in its preceding convolution is possible because batch normalization is an affine transformation. However, in BNNs the batch normalization layer cannot be absorbed in the preceding convolution as it would require its weights to have much higher precision. Thus, in BNNs the batch normalization layer is fused into the proceeding sign-function. The fused operator is illustrated by:
\begin{equation}
    \text{sign-batchnorm}(Y) = \begin{cases}
    +1, & \text{if } \hat{Y} \geq 0 \\
    -1, & elsewhere
    \end{cases}
\end{equation}
where $Y$ is the output of a binary convolution $(X_B * W_B)$. $\hat{Y}$ can be represented as a function of $Y$ as follows:
\begin{eqnarray}
    \hat{Y} \equiv \gamma \frac{Y-\mu}{\sqrt{\sigma^2 + \epsilon}} + \beta \geq 0 \\
    \gamma(Y-\mu) \geq -\beta\sqrt{\sigma^2 + \epsilon} \\
    \begin{cases}\label{eq:signbn2case}
        Y \geq \mu - \frac{\beta\sqrt{\sigma^2 + \epsilon}}{\gamma} \text{, if } \gamma > 0 \\
        Y \leq \mu - \frac{\beta\sqrt{\sigma^2 + \epsilon}}{\gamma} \text{, if } \gamma < 0
    \end{cases}
\end{eqnarray}
Eq. \ref{eq:signbn2case} can be further simplified by fusing parts into the preceding convolution, which eliminates the need for both a greater-than and lower-than comparison. Since both $\gamma$ and the weights ($W_B$) are feature map-based, the weights' feature maps for which $\gamma$ is negative can be multiplied by -1 such that the following equation remains:
\begin{equation} \label{eq:binconv}
    \text{sign-batchnorm}(X_B * \pm W_B) \geq \pm\mu - \frac{\beta\sqrt{\sigma^2 + \epsilon}}{\mid \gamma \mid}
\end{equation}
where $X_B$ is the input to the binary convolution. 

\subsection{Training} \label{sec:background:training}
\begin{figure}
  \centering
  \includegraphics[width=0.9\textwidth]{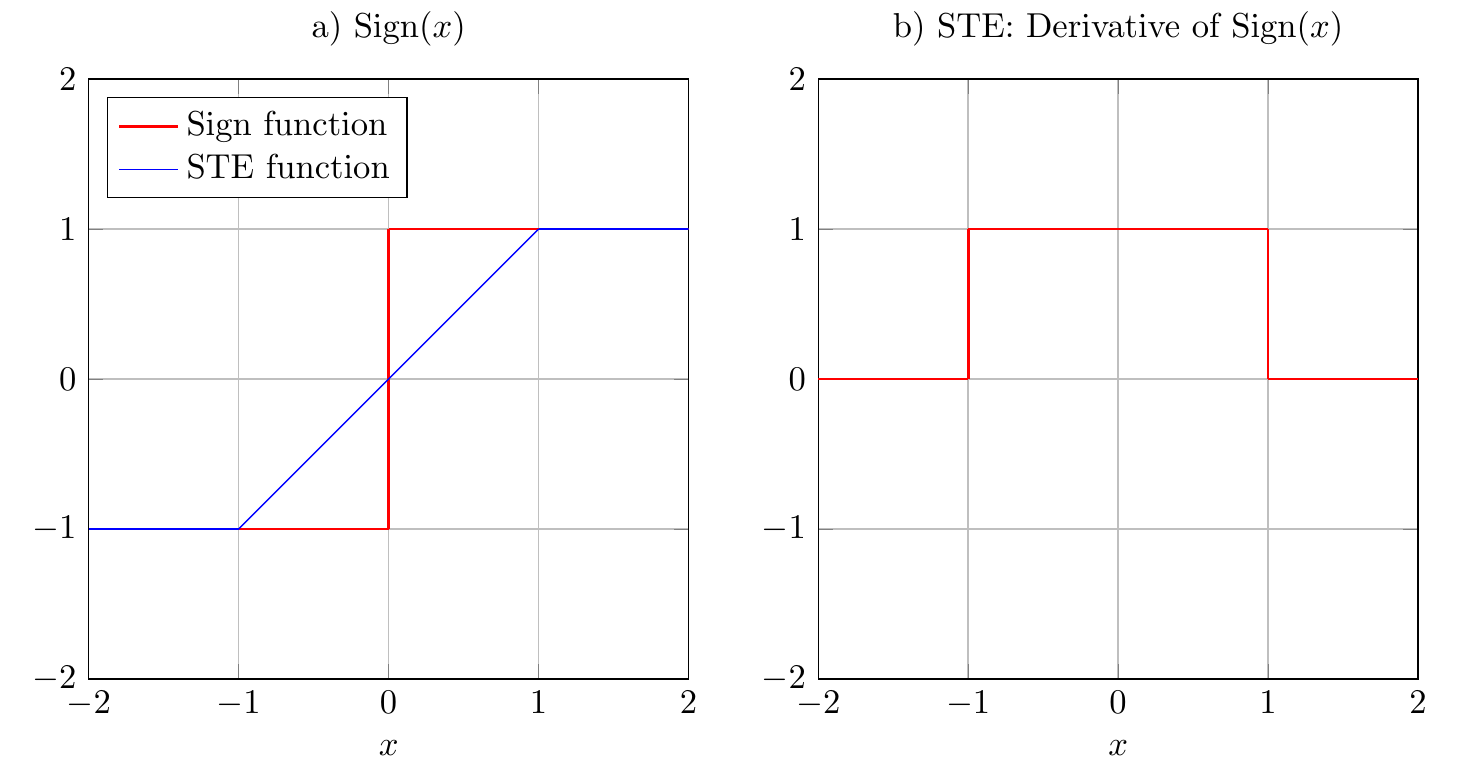}
  \caption{The $\text{sign}(x)$ function and the use of a STE to enable gradients.}
  \label{fig:ste}
\end{figure}
Aside from the efficient inference, the binary weights of a BNN still need to be learned as well. Similar to a real-valued CNN it is possible to use the gradient descent algorithm. However, since the sign-function (equation \ref{eqn:sign}) has a derivative that is almost everywhere $0$, it is not possible to train BNNs directly using the famous gradient descent algorithm. Fortunately, this issue has been resolved by the introduction of a so-called straight-through estimator (STE) \cite{Tieleman2012COURSERA:Learning}. The STE is illustrated in Fig. \ref{fig:ste} and can be expressed as a clipped identity function:
\begin{equation}
    \frac{\partial X_B}{\partial X_R} = 1_{|X_R \le 1|}
\end{equation}
where $X_R$ is the real-valued input to the sign-function, $X_B$ the binarized output and $1_{|X_R \le 1|}$ evaluates to $1$ if $|X_R| \le 1$ and $0$ otherwise (Fig. \ref{fig:ste}b). As such, the complete gradient chain of the loss with respect to the real-valued non-quantized weights is as follows:
\begin{equation}
    \frac{\partial L}{\partial X_R} = \frac{\partial L}{\partial X_B} \frac{\partial X_B}{\partial X_R} = \frac{\partial L}{\partial X_B} * 1_{|X_R \le 1|}
\end{equation}
where $L$ is the loss function, $X_R$ is the real-valued input to the sign-function, $X_b$ is the binarized output and $1_{|X_R \le 1|}$ is the clipped identity function. With the help of an STE BNNs can be trained using the same gradient descent algorithms as in ordinary real-valued CNNs. However, note that even with the use of STEs BNNs cannot be trained up to satisfactory performance and a substantial accuracy gap with respect to their full-precision counterparts remains.

\section{Classification of accuracy repair techniques} \label{sec:classification}
Although the premise of BNNs with high compression and efficient compute logic is good, the accuracy loss gotten by transforming a real-valued CNN into a BNN is not acceptable. On one side the accuracy loss is due to the large quantization error between full-precision and binary values. On the other side, this is due to the discrete optimization landscape for which gradient descent-based optimization has difficulties, even with the use of STEs. Thus, BNNs heavily suffer from accuracy loss. 

Fortunately, in recent years many training techniques and network topology changes have been proposed that aim to reduce the accuracy loss. Whereas training techniques aim to improve the optimization process of BNNs, network topology changes aim to reduce the quantization error between full-precision CNNs and their binary counterpart.

\begin{table}[]
\caption{Classification table of accuracy repair methods and legend for table 2.}
\label{tab:classification}
\begin{tabular}{@{}llll@{}}
\toprule
 & \textbf{Property} & \textbf{Abbreviation} & \textbf{Description} \\ \midrule
\multirow{22}{*}{\textbf{\rotatebox[origin=c]{90}{Training technique}}} & \multirow{8}{*}{1. Binarizer (STE)} & LC\_$|X|$ & Linear clipped at $|X|$ \\
 &  & LC\_A & Linear with adaptive clipping \\
 &  & PN & Polynomial \\
 &  & GPN & Gradual Polynomial \\
 &  & T & Gradual tanh-based sign \\
 &  & EDE & T with magnitude scaling \\
 &  & SS & SwishSign \\
 &  & EWGS & Element-wise Gradient Scaling \\ \cmidrule(l){2-4} 
 & \multirow{6}{*}{2. Normalization} & LB & Learnable bias \\
 &  & DB & Dynamic bias \\
 &  & STD & Division by standard deviation \\
 &  & MSTD & Zero-mean and division by standard deviation \\
 &  & MSTDB & MSTD divided by $b$ \\
 &  & BN & Batch normalization \\ \cmidrule(l){2-4} 
 & \multirow{2}{*}{3. Teacher-Student} & TO & Train towards Teacher's output \\
 &  & TB & TO and per-block loss with Teacher \\ \cmidrule(l){2-4} 
 & \multirow{2}{*}{4. Regularization} & RE & Weight entropy regularization \\
 &  & RD & Weight distance to 1 regularization \\ \cmidrule(l){2-4} 
 & 5. Two-stage training & TST\_Y/N & Yes / No \\ \cmidrule(l){2-4} 
 & \multirow{3}{*}{6. Optimizer} & SGD & SGD optimizer \\
 &  & ADAM & ADAM optimizer \\
 &  & Custom & Custom BNN optimizer \\ \midrule
\multirow{13}{*}{\textbf{\rotatebox[origin=c]{90}{Topology changing}}} & \multirow{3}{*}{7. Scaling factor} & AM & Absolute mean of the Weights \\
 &  & LF & Learnable factor \\
 &  & LFI & Learnable factor with initialization \\ \cmidrule(l){2-4} 
 & \multirow{3}{*}{8. Ensemble} & EL & Per-layer \\
 &  & EB & Per-block \\
 &  & EN & Per-network \\ \cmidrule(l){2-4} 
 & \multirow{5}{*}{9. Activation function} & I\&H & Identity and htanh \\
 &  & ReLU & ReLU \\
 &  & PReLU & PReLU \\
 &  & RPReLU & RPReLU \\
 &  & DPReLU & DPReLU \\ \cmidrule(l){2-4} 
 & 10. Double residual & 2R\_Y/N & Residual per convolution Yes / No \\ \cmidrule(l){2-4} 
 & 11. Squeeze-and-excitation & SE\_Y/N & Squeeze-and-Excitation Yes / No \\ \bottomrule
\end{tabular}
\end{table}

As such, in this section, we employ a hierarchical classification of accuracy repair methods, as illustrated in Table \ref{tab:classification}. This classification starts with two main branches: training techniques and network topology changes. Within these branches are several categories that group accuracy repair techniques together. Note that the repair categories can be applied orthogonally to each other, but repairs within a single category cannot (except for some of the teacher-student and regularization approaches).

Within the \textbf{training techniques branch} there are the following six categories:
\begin{enumerate}
    \item \textit{Binarizer (STE)} - Since the derivative of the binarization process (sign-function) is 0 almost everywhere, works have come up with various solutions to create a fake derivative and thereby enable the use of the renowned backpropagation algorithm.
    \item \textit{Normalization} - Before binarizing the features or weights, some works apply a form of normalization. Normalization includes, but is not limited to, standardizing and/or centralizing. Moreover, since normalization is proceeded by a binarizer, any multiplication done in the normalization phase will only affect the backward pass, i.e. gradients will be scaled.
    \item \textit{Teacher-Student} - The Teacher-Student methodology aims to help train CNNs by having a complex teacher network trying to transfer knowledge to a simple student network (the BNN). The knowledge transfer process can be done at distinct stages in the network.
    \item \textit{Regularization} - Regularization is a technique that introduces another loss term that should indirectly help train the network. The final network loss $L$ can then be described as:
    \begin{equation}
        L=L_{CE}+\lambda L_{R}
    \end{equation}
    where $L_{CE}$ the cross-entropy loss, $L_{R}$ the regularization loss, and $\lambda$ the balancing coefficient.
    \item \textit{Two-stage training} - The default in quantization-aware training is to train with both quantized features and weights simultaneously. In the two-stage training procedure, the first stage employs only binary features, whereas in the second stage both are binary.
    \item \textit{Optimizer} - Within BNN works some use the SGD optimizer, whereas others use the ADAM optimizer. Contrary to real-valued networks in which SGD supersedes ADAM in terms of accuracy, it is not clear if this is the case for BNNs.
\end{enumerate}

Within the \textbf{network topology changing branch} there are the following five categories:
\begin{enumerate}
    \setcounter{enumi}{6}
    \item \textit{Scaling factor} - Usually a binary convolution outputs values with a larger range than its real-valued counterpart. Hence, many works employ a scaling factor $\alpha$ to minimize the quantization error.
    \begin{equation}
        \min\limits_{\alpha} \mid Y - \alpha Y_B \mid^2
    \end{equation}
    where $Y_B$ is the output of a binary-valued convolution and $Y$ is its real-valued counterpart.
    \item \textit{Ensemble} - Ensemble methods use multiple instantiations of (a certain part of) the network to obtain better performance than could be obtained from any individual part.
    \item \textit{Activation function} - Next to the binarizer that can be regarded as the non-linearity in the network, works have suggested adding another activation function, such as ReLU or PReLU, to improve the accuracy.
    \item \textit{Double residual} - To keep the flow of information rich in the network, BNNs use a residual connection after every convolution, instead of after multiple convolutions. 
    \item \textit{Squeeze-and-Excitation} - The concept of Squeeze-and-Excitation is to assign each output feature map channel a different weight based on its importance. These channel-wise weights are based on the input feature maps.
\end{enumerate}

These eleven categories are used to create an overview of the repair methods used by BNN papers, which is presented in Section \ref{sec:overview}.

\section{Overview of accuracy repair techniques as applied in the literature} \label{sec:overview}
Accuracy repair techniques can be grouped into two main branches: training techniques and network topology changes. These two branches can be further split up and classified according to Table \ref{tab:classification}. Table \ref{tab:overview} presents an overview of most BNN works and their respective repair techniques used. Additionally, their claimed top-1 accuracies on the ResNet-18 architecture and ImageNet dataset are shown. As a reference, the first row in Table \ref{tab:overview} denotes the full-precision ResNet-18 architecture with its accuracy on ImageNet.

\begin{table}[]
\caption{BNN training techniques classification and evaluation. Refer to table 1 for a legend.}
\label{tab:overview}
\resizebox{\textwidth}{!}{%
\begin{tabular}{@{}llllllllllllllll@{}}
\toprule
\textbf{Work} & \textbf{Year} & \textbf{\rotatebox[origin=c]{90}{\begin{tabular}[c]{@{}l@{}}ResNet-18 \&\\ ImageNet\\ Accuracy (\%)\end{tabular}}} & \textbf{\rotatebox[origin=c]{90}{\begin{tabular}[c]{@{}l@{}}1. Feature\\ Binarizer\end{tabular}}} & \textbf{\rotatebox[origin=c]{90}{\begin{tabular}[c]{@{}l@{}}2. Feature \\ Normalization\end{tabular}}} & \textbf{\rotatebox[origin=c]{90}{\begin{tabular}[c]{@{}l@{}}1. Weight \\ Binarizer\end{tabular}}} & \textbf{\rotatebox[origin=c]{90}{\begin{tabular}[c]{@{}l@{}}2. Weight \\ Normalization\end{tabular}}} & \textbf{\rotatebox[origin=c]{90}{3. Teacher-Student}} & \textbf{\rotatebox[origin=c]{90}{4. Regularization}} & \textbf{\rotatebox[origin=c]{90}{5. Two-stage   training}} & \textbf{\rotatebox[origin=c]{90}{6. Optimzier}} & \textbf{\rotatebox[origin=c]{90}{7. Scaling   factor}} & \textbf{\rotatebox[origin=c]{90}{8. Ensemble}} & \textbf{\rotatebox[origin=c]{90}{9. Activation function}} & \textbf{\rotatebox[origin=c]{90}{10. Double residual}} & \textbf{\rotatebox[origin=c]{90}{11. Squeeze-and-Excitation}} \\ \midrule
Full-precision   ResNet \cite{He2015DeepRecognition} & \citeyear{He2015DeepRecognition} & 69.3 &  &  & \textbf{} & \textbf{} & \textbf{} & \textbf{} & \textbf{} & \textbf{} & \textbf{} & \textbf{} & \textbf{} & \textbf{} & \textbf{} \\
BinaryNet \cite{Courbariaux2016Binarized-1} & \citeyear{Courbariaux2016Binarized-1} & 42.2 & LC\_1 &  & LC\_1 &  &  &  & N & SGD &  &  & I\&H & N & N \\
XNORNet   \cite{Rastegari2016XNOR-Net:Networks} & \citeyear{Rastegari2016XNOR-Net:Networks} & 51.2 & LC\_1 & BN & LC\_1 &  &  &  & N & ADAM & AM &  & I\&H & N & N \\
ABC-Net (5x5)   \cite{Lin2017TowardsNetwork} & \citeyear{Lin2017TowardsNetwork} & 65 & LC\_1 & DB & LC\_1 & DB &  &  & N & ADAM & LF & EL & I\&H & N & N \\
Compact BNN   \cite{Tang2017HowAccuracy} & \citeyear{Tang2017HowAccuracy} & n/a & LC\_1 &  & LC\_1 &  &  &  & N & ADAM &  &  & PReLU & N & N \\
Regularized   BNN \cite{Darabi2018RegularizedTraining} & \citeyear{Darabi2018RegularizedTraining} & 53 & SS &  & SS &  &  & RE & N & n/a & AM &  & n/a & N & N \\
Bi-Real Net   \cite{Liu2018Bi-RealAlgorithm} & \citeyear{Liu2018Bi-RealAlgorithm} & 56.4 & PN &  & LC\_1 &  &  &  & N & ADAM & AM &  & None & Y & N \\
Continuous   Binarization \cite{Sakr2018TrueBinarization} & \citeyear{Sakr2018TrueBinarization} & n/a & LC\_A &  & n/a &  &  & $M^1$ & N & SGD & LF &  & n/a & N & N \\
BinaryDenseNet   \cite{Bethge2019BinaryDenseNet:Networks} & \citeyear{Bethge2019BinaryDenseNet:Networks} & n/a & LC\_1.3 &  & LC\_1.3 &  &  &  & N & ADAM &  &  & ReLU & $M^1$ & N \\
Group-Net (5x)   \cite{Zhuang2019StructuredSegmentation} & \citeyear{Zhuang2019StructuredSegmentation} & 67 & PN &  & LC\_$\infty$ &  &  &  & N & SGD & AM & EB & ReLU & Y & N \\
Human pose   estimation \cite{Bulat2019ImprovedRecognition} & \citeyear{Bulat2019ImprovedRecognition} & 53.7 & T & BN & T &  & TO &  & Y & ADAM & AM &  & PReLU & $M^1$ & N \\
XNORNet++   \cite{Bulat2019XNOR-Net++:Networks} & \citeyear{Bulat2019XNOR-Net++:Networks} & 57.1 & n/a & BN & n/a &  &  &  & N & ADAM & LF &  & n/a & N & N \\
Latent Weights   \cite{Helwegen2019LatentOptimization} & \citeyear{Helwegen2019LatentOptimization} & n/a &  &  &  &  &  &  &  & Custom &  &  &  &  &  \\
BENN (6x)   \cite{Zhu2019BinaryBit} & \citeyear{Zhu2019BinaryBit} & 61 & LC\_1 &  & LC\_1 &  &  &  & N & ADAM &  & EN & n/a & N & N \\
Circulant BNN   \cite{Liu2019CirculantPropagation} & \citeyear{Liu2019CirculantPropagation} & 61.4 & PN & BN & PN &  &  &  & N & SGD & AM &  & PReLU & Y & N \\
CI-BCNN   \cite{Wang2019LearningNetworks} & \citeyear{Wang2019LearningNetworks} & 59.9 & LC\_1 &  & LC\_1 &  &  &  & N & SGD &  &  & I\&H & N & N \\
IR-Net   \cite{Qin2020ForwardNetworks} & \citeyear{Qin2020ForwardNetworks} & 58.1 & EDE &  & EDE & MSTD &  &  & N & SGD & AM &  & I\&H & Y & N \\
BBG   \cite{Shen2020BalancedResidual} & \citeyear{Shen2020BalancedResidual} & 59.4 & LC\_1 &  & LC\_1 & $M^1$ &  &  & N & n/a & AM &  & n/a & Y & N \\
Real-to-Binary   \cite{Martinez2020TrainingConvolutions} & \citeyear{Martinez2020TrainingConvolutions} & 65.4 & LC\_1 & BN & LC\_1 &  & TB &  & Y & ADAM & LF &  & PReLU & Y & N \\
ReActNet   \cite{Liu2020ReActNet:Functions} & \citeyear{Liu2020ReActNet:Functions} & 65.9 & PN & LB & LC\_1 &  & TO &  & Y & ADAM & AM &  & PReLU & Y & N \\
RBNN   \cite{Lin2020RotatedNetwork} & \citeyear{Lin2020RotatedNetwork} & 59.9 & GPN & MSTD & GPN & MSTD &  &  & N & SGD & LFI &  & I\&H & Y & N \\
ProxyBNN   \cite{He2020ProxyBNN:Matrices} & \citeyear{He2020ProxyBNN:Matrices} & 63.7 & LC\_1 & BN & $M^1$ &  &  &  & N & ADAM & AM &  & PReLU & Y & N \\
Noisy   Supervision \cite{Han2020TrainingSupervision} & \citeyear{Han2020TrainingSupervision} & 59.4 & LC\_1 &  & LC\_1 &  &  & $M^1$ & N & SGD & AM &  & PReLU & Y & N \\
Information in   BNN \cite{Ignatov2020ControllingNetwork} & \citeyear{Ignatov2020ControllingNetwork} & 61.36 & $M^1$ &  & T &  &  &  & N & n/a &  &  & n/a & N & N \\
BNN SISR   \cite{Xin2020BinarizedResolution} & \citeyear{Xin2020BinarizedResolution} & n/a & LC\_1 & BN & LC\_1 &  &  &  & N & ADAM & AM &  & n/a & Y & N \\
SI-BNN   \cite{Wang2020Sparsity-InducingNetworks} & \citeyear{Wang2020Sparsity-InducingNetworks} & 59.7 & $M^1$ &  & LC\_1 &  &  &  & N & ADAM & AM &  & None & Y & N \\
Adam \&   TST \cite{Liu2021HowOptimization} & \citeyear{Liu2021HowOptimization} & n/a & PN & LB & LC\_1 &  & TO &  & Y & ADAM & AM &  & PReLU & Y & N \\
ReCU   \cite{Xu2021ReCU:Networks} & \citeyear{Xu2021ReCU:Networks} & 66.4 & PN & STD & LC\_A & MSTDB &  &  & N & SGD & LF &  & PReLU & Y & N \\
Bop and beyond   \cite{Suarez-Ramirez2021ANetworks} & \citeyear{Suarez-Ramirez2021ANetworks} & n/a &  &  &  &  &  &  &  & Custom &  &  &  &  &  \\
Sub-bit BNN   \cite{Wang2021Sub-bitNetworks} & \citeyear{Wang2021Sub-bitNetworks} & 55.7 & PN & LB & LC\_1 &  &  &  & Y & ADAM & AM &  & PReLU & Y & N \\
MeliusNet   \cite{Bethge2021MeliusNet:Networks} & \citeyear{Bethge2021MeliusNet:Networks} & n/a & LC\_1.3 & BN & LC\_1.3 &  &  &  & N & RADAM &  &  & ReLU & $M^1$ & N \\
Increasing   Entropy \cite{Zou2021IncreasingNetworks} & \citeyear{Zou2021IncreasingNetworks} & 58.5 & EDE & $M^1$ & EDE & $M^1$ &  & $M^1$ & N & SGD & AM &  & I\&H & Y & N \\
SA-BNN   \cite{Liu2021SA-BNN:Network} & \citeyear{Liu2021SA-BNN:Network} & 61.7 & PN & BN & LC\_1 &  &  &  & N & ADAM & LF &  &  & Y & N \\
UaBNN   \cite{Zhao2021Uncertainty-awareNetworks} & \citeyear{Zhao2021Uncertainty-awareNetworks} & 61.9 & PN & BN & $M^1$ &  &  &  & N & ADAM & AM &  & RPReLU & Y & N \\
Lottery ticket   BNN \cite{Diffenderfer2021Multi-PrizeNetwork} & \citeyear{Diffenderfer2021Multi-PrizeNetwork} & n/a & PN &  & LC\_1 &  &  &  & N & SGD & AM &  & ReLU & N & N \\
Expert BNN   \cite{Bulat2021High-CapacityNetworks} & \citeyear{Bulat2021High-CapacityNetworks} & n/a & LC\_1 & BN & LC\_1 &  & TB &  & Y & ADAM & LF &  & PReLU & Y & N \\
BNN-BN   \cite{Chen2021BNNNormalization} & \citeyear{Chen2021BNNNormalization} & 61.1 & PN & LB & LC\_1 & MSTDB & TO &  & Y & ADAM & AM &  & PReLU & Y & N \\
BCNN   \cite{Redfern2021BCNN:Precision} & \citeyear{Redfern2021BCNN:Precision} & n/a & PN & LB & LC\_1 &  &  &  & Y & ADAM &  & EB & PReLU & Y & N \\
BNN-BN SR   \cite{Jiang2021TrainingSuper-Resolution} & \citeyear{Jiang2021TrainingSuper-Resolution} & n/a & PN &  & LC\_1 &  & TB &  & N & ADAM &  &  & RPReLU & N & N \\
Complex BNN   \cite{Li2021BCNN:Network} & \citeyear{Li2021BCNN:Network} & 57.65 & LC\_1 &  & LC\_1 &  &  &  & N & ADAM &  &  & I\&H & Y & N \\
EWGS   \cite{Lee2021NetworkScaling} & \citeyear{Lee2021NetworkScaling} & 55.3 & EWGS &  & EWGS &  &  &  & N & ADAM & LFI &  & ReLU & N & N \\
Equal Bits   \cite{Li2021EqualWeights} & \citeyear{Li2021EqualWeights} & 60.4 & $M^1$ & DB & $M^1$ & DB &  &  & N & SGD &  &  & n/a & n/a & N \\
DyBNN   \cite{Zhang2021DynamicThresholds} & \citeyear{Zhang2021DynamicThresholds} & 67.4 & n/a & DB & n/a &  & TO &  & Y & ADAM &  &  & PReLU & Y & Y \\
SD-BNN   \cite{Xue2021Self-DistributionNetworks} & \citeyear{Xue2021Self-DistributionNetworks} & 66.5 & EDE & DB & EDE & DB &  &  & Y & SGD &  &  & PReLU & Y & Y \\
BNN latent   weights \cite{Xu2021ImprovingWeights} & \citeyear{Xu2021ImprovingWeights} & 63.8 & LC\_1 & BN & LC\_1 &  & TO & $M^1$ & N & ADAM & AM &  & PReLU & Y & N \\
PokeBNN   \cite{Zhang2021PokeBNN:Accuracy} & \citeyear{Zhang2021PokeBNN:Accuracy} & n/a & LC\_3 &  & LC\_1 &  &  &  & N & ADAM &  &  & DPReLU & Y & Y \\
BoolNet   \cite{Guo2021BoolNet:Networks} & \citeyear{Guo2021BoolNet:Networks} & n/a & LC\_1 & BN & T &  & TO &  & N & RADAM &  &  & None & $M^1$ & N \\ \midrule
\multicolumn{16}{l}{$M^1$denotes miscellaneous methods}  \\ \bottomrule
\end{tabular}%
}
\end{table}

\subsection{Training techniques}
The training technique branch can be split up into the following categories: binarizer, normalization, teacher-student, regularization, two-stage training, and optimizer. Each repair method within these categories is outlined in the sections below. Note that these repair methods only influence the training phase, not the inference phase.

\subsubsection{Binarizer (STE)}
Similar to real-valued CNNs training of BNNs is done using the backpropagation algorithm. This algorithm requires every function to be differentiable. Since the derivative of the binarizer (sign-function) is 0 almost everywhere, \textbf{\citeauthor{Tieleman2012COURSERA:Learning} (\citeyear{Tieleman2012COURSERA:Learning})} \cite{Tieleman2012COURSERA:Learning} introduced the default STE (Fig. \ref{fig:ste}). It passes gradients as is for values that fall in the interval $[-1, +1]$, but cancels gradients for values that are outside the interval. An STE is often used to provide an estimate for the gradient. Obviously, this introduces a gradient mismatch between the actual gradient of the sign-function and the STE, which makes the training of BNNs more difficult. Widely explored are alternative estimators for the gradient, which should aid the optimization of a BNN.


In recent years, there have been works that change the clipping interval of the STE. For example, BinaryDenseNet \cite{Bethge2019BinaryDenseNet:Networks} and MeliusNet \cite{Bethge2021MeliusNet:Networks} use an interval of $[-1.3, +1.3]$, whereas PokeBNN \cite{Zhang2021PokeBNN:Accuracy} uses an interval of $[-3, +3]$. These papers chose this interval based on empirical studies. Next to a fixed interval, \cite{Sakr2018TrueBinarization} proposes to make the interval a learnable parameter. They constrain the parameter to be small using an additional loss term. Likewise ReCU \cite{Xu2021ReCU:Networks} changes the interval during training from $[-0.85, +0.85]$ to $[-0.99, +0.99]$. According to their mathematics, the smaller interval is best for reducing the quantization error, whereas the higher interval ensures maximum information entropy.

\begin{figure}
  \centering
  \includegraphics[width=0.9\textwidth]{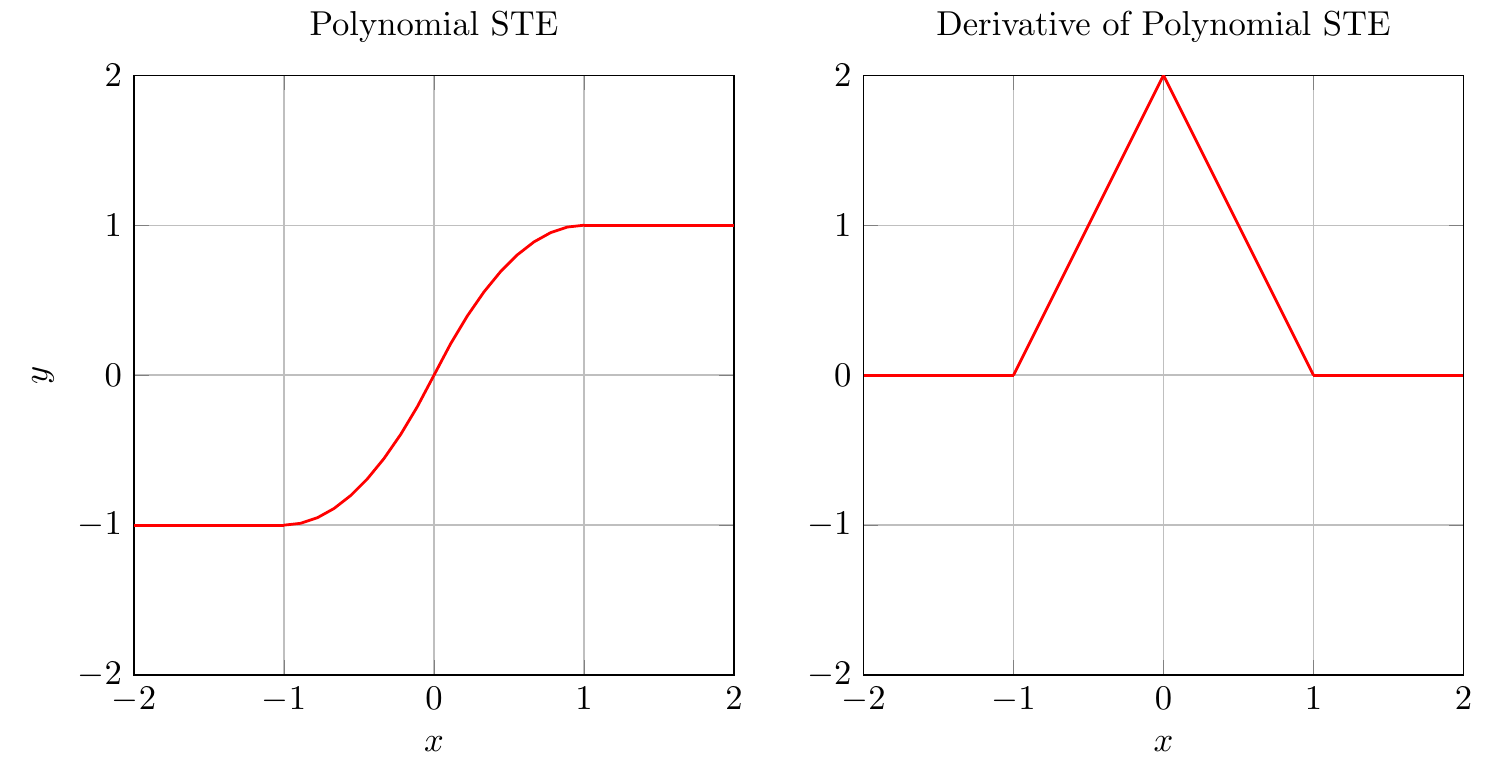}
  \caption{The polynomial STE function and its derivative.}
  \label{fig:ste_overview_pn}
\end{figure}
\textbf{Bi-Real Net (\citeyear{Liu2018Bi-RealAlgorithm})} \cite{Liu2018Bi-RealAlgorithm} introduced the Polynomial STE, shown in Fig. \ref{fig:ste_overview_pn}. The triangular-shape of the derivative should resemble the actual derivative of a sign-function more closely and therefore the gradient approximation error would be less. As can be seen in Table \ref{tab:overview} many works have adapted this STE for feature binarization.

\begin{figure}
  \centering
  \includegraphics[width=0.9\textwidth]{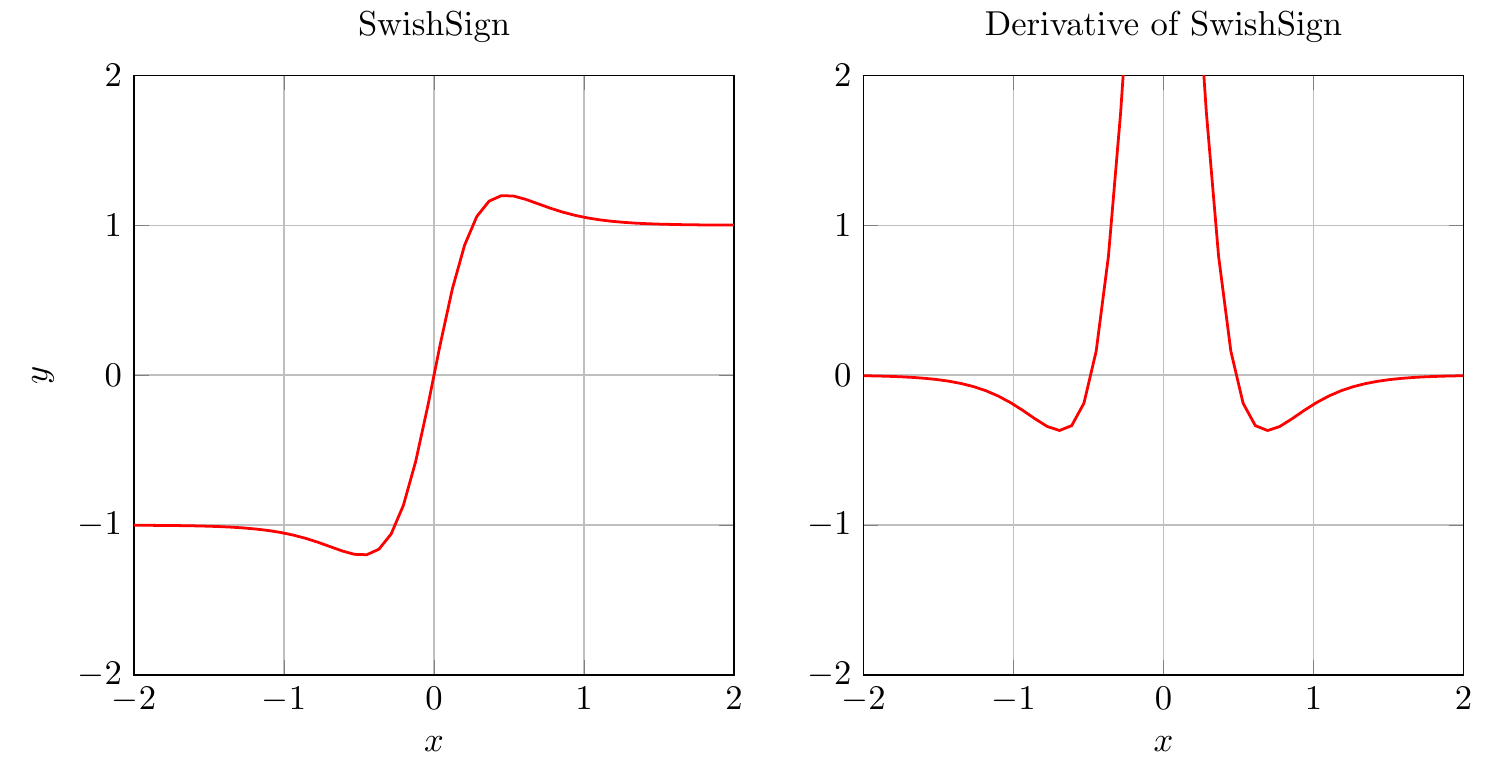}
  \caption{The SwishSign function and its derivative.}
  \label{fig:ste_overview_ss}
\end{figure}
\textbf{Regularized BNN (\citeyear{Darabi2018RegularizedTraining})} \cite{Darabi2018RegularizedTraining} introduced the SwishSign STE, illustrated in Fig. \ref{fig:ste_overview_ss}. It is based on the sigmoid-function. Similar to the polynomial STE, this function should resemble the actual derivative of a sign-function more closely. However, this derivative includes certain intervals for which the gradient is negative.
The equation corresponding to the SwishSign is:
\begin{equation}
    SS_{\beta}(x) = 2 \sigma (\beta x) (1 + \beta x (1 - \sigma (\beta x))) - 1
\end{equation}
where $\sigma$ represents the sigmoid function and $\beta$ equals 5 as in Fig. \ref{fig:ste_overview_ss}. 

\begin{figure}
  \centering
  \includegraphics[width=0.9\textwidth]{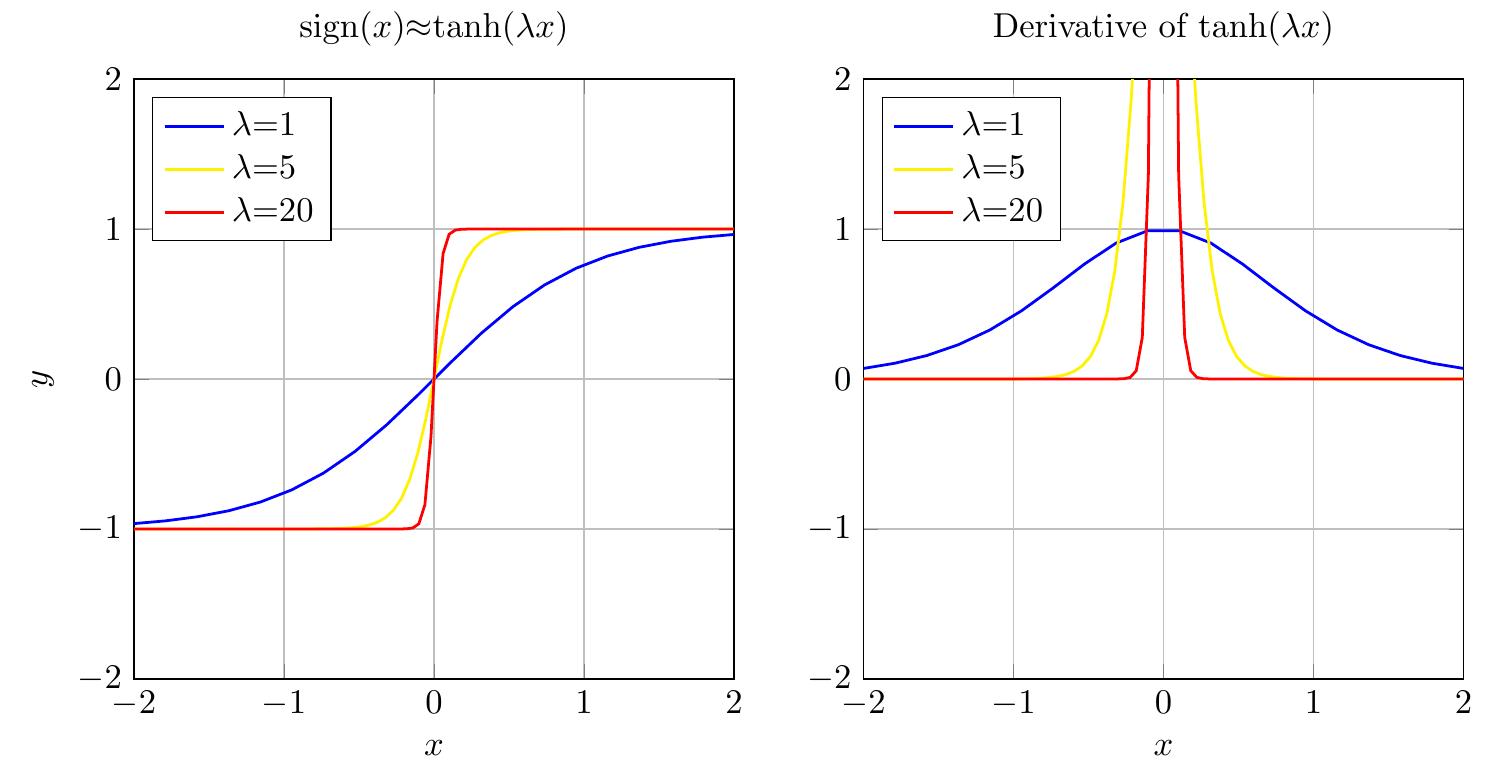}
  \caption{The gradual tanh($\lambda x$) function, in which $\lambda$ is changed during training.}
  \label{fig:ste_overview_tanh}
\end{figure}
\textbf{Human pose estimation BNN (\citeyear{Bulat2019ImprovedRecognition})} \cite{Bulat2019ImprovedRecognition} introduced a STE that is gradually changing during the training process. It is based on the tanh function, shown in Fig. \ref{fig:ste_overview_tanh}, and approximates the actual gradient of a sign-function more and more as $\lambda$ is increased.

\begin{figure}
  \centering
  \includegraphics[width=0.9\textwidth]{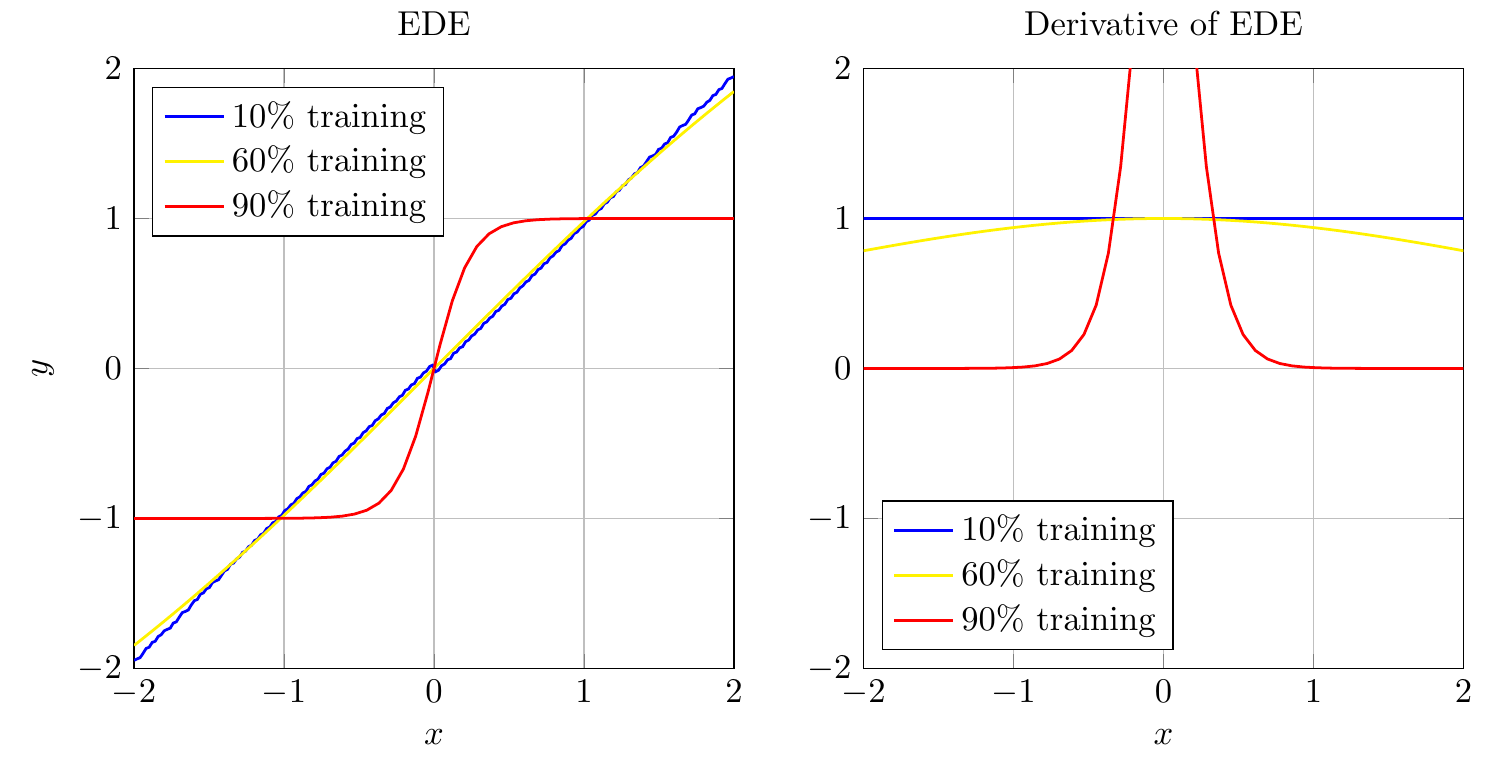}
  \caption{The EDE function that is gradually altered based on the stage of training.}
  \label{fig:ste_overview_ede}
\end{figure}
\textbf{IR-Net (\citeyear{Qin2020ForwardNetworks})} \cite{Qin2020ForwardNetworks} extended the approach of \cite{Bulat2019ImprovedRecognition} by scaling the magnitude accordingly and calls its STE EDE. EDE can be described by:
\begin{eqnarray}
    \text{EDE}(x) = \text{max}(\frac{1}{\lambda}, 1) * \text{tanh}(\lambda * x) \\
    \lambda = 10^{-3 + (1+3) * T}
\end{eqnarray}
where $T$ is the percentage of training. This extension is shown in Fig. \ref{fig:ste_overview_ede}. 

\begin{figure}
  \centering
  \includegraphics[width=0.9\textwidth]{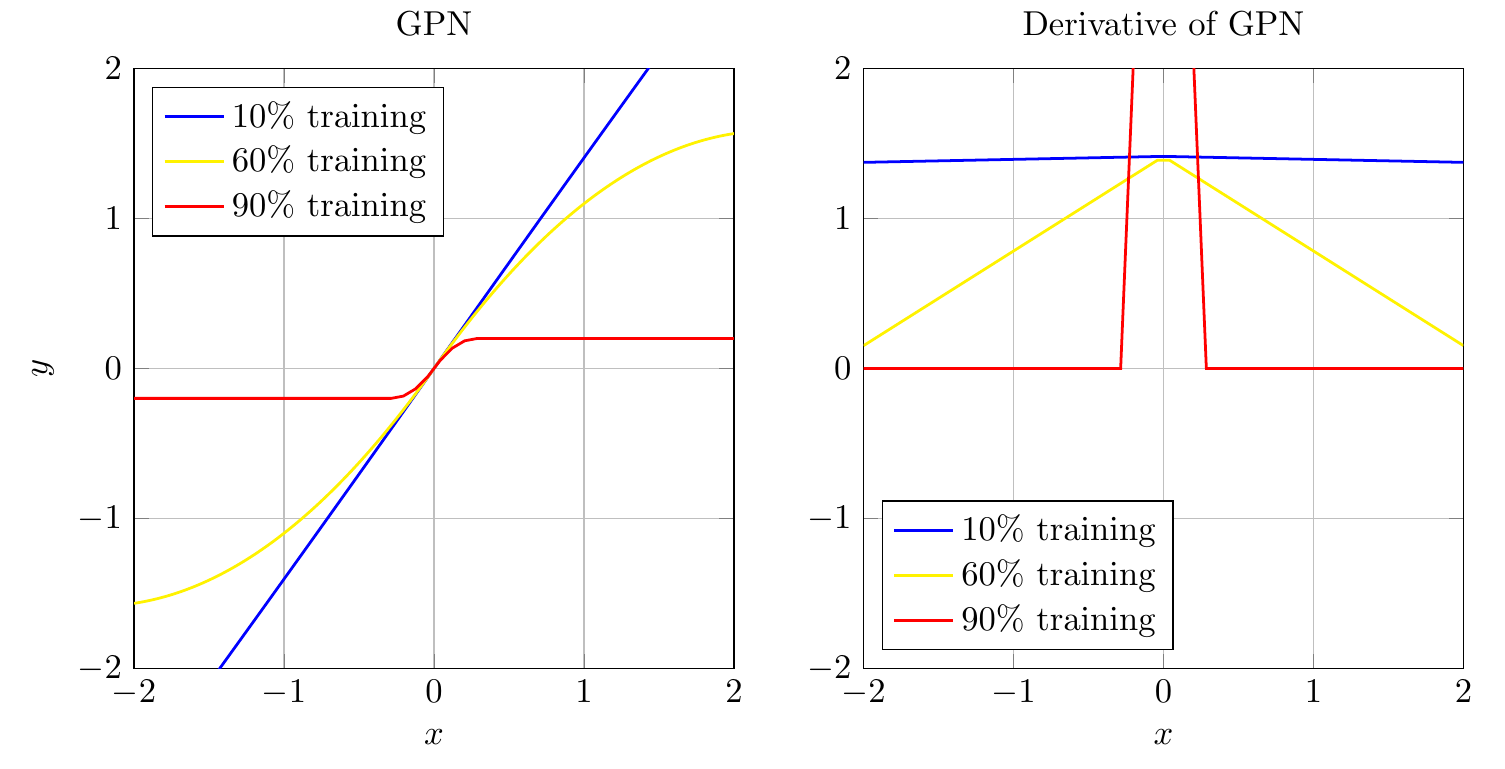}
  \caption{The GPN function that is gradually altered based on the stage of training.}
  \label{fig:ste_overview_gpn}
\end{figure}
\textbf{RBNN (\citeyear{Lin2020RotatedNetwork})} \cite{Lin2020RotatedNetwork} is a variant upon the polynomial STE: GPN. The triangular shape of the polynominial is gradually sharpened and compressed based on the stage of training. It is illustrated in Fig. \ref{fig:ste_overview_gpn} and its equation is:
\begin{eqnarray}
    \text{GPN}(x) = \begin{cases}
        k \cdot (-\text{sign}(x) \frac{\lambda^2 x^2}{2} + \sqrt{2}\lambda x) , & \text{if } \mid x \mid  < \frac{\sqrt{2}}{\lambda} \\
        k \cdot \text{sign}(x), & \text{otherwise}
    \end{cases} \\
    \lambda = 10^{-2 + (1+2) * T} \\
    k = \text{max}(1/\lambda, 0)
\end{eqnarray}
where $T$ is the percentage of training. 

\begin{figure}
  \centering
  \includegraphics[width=\textwidth]{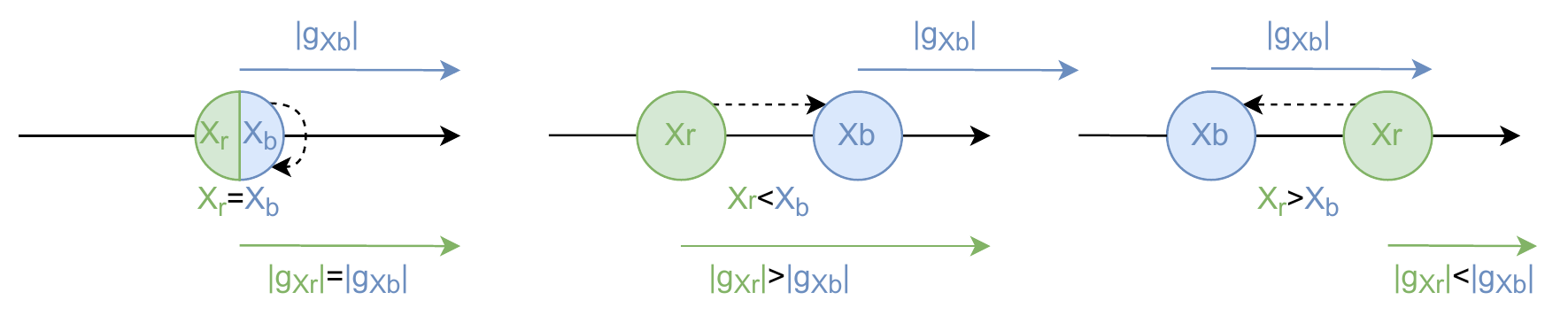}
  \caption{Example of EWGS approach, corresponding to eq. \ref{eq:ewgs}. Dashed line indicates binarization. Three distinct cases are shown, in which the real-value $x_r$ is equal (left), smaller than (middle), or larger than (right) the binary value $x_b$. Depending on the difference between the real and binary value, the gradient of the real value $g_{x_r}$ is scaled up, scaled down, or equal to the gradient of the binary value $g_{x_b}$.}
  \label{fig:ewgs}
\end{figure}

\textbf{EWGS (\citeyear{Lee2021NetworkScaling})} \cite{Lee2021NetworkScaling} notices that the default STE has two issues: 1. Multiple pre-quantized values can produce the same discrete quantized value, 2. The gradient provided by the discrete value affects each of the latent values differently. To solve these issues, EWGS suggests the following alternative:
\begin{equation}\label{eq:ewgs}
    g_{x_r} \approx g_{x_b} (1 + \delta \text{sign}(g_{x_b})(x_r-x_b))
\end{equation}
where $g_{x_r}$ and $g_{x_b}$ are the partial derivatives of the loss with respect to $x_r$ and $x_b$, and $\delta \geq 0 $ a scaling factor. This approach is illustrated in Fig. \ref{fig:ewgs}. Based on the scaling factor, the sign of the element and the quantization error ($x_r-x_b$) EWGS adaptively adjusts the gradient. $\delta$ can be a fixed value, e.g. $1e-3$, or the second-order derivative of the loss with respect to the discrete value $x_b$.

\subsubsection{Normalization}
Before binarization of features or weights, some BNN works apply a form of normalization. Normalization might shift and/or scale weights and/or features.

\textbf{XNOR-Net (\citeyear{Rastegari2016XNOR-Net:Networks})} \cite{Rastegari2016XNOR-Net:Networks} used a batch normalization (BN) layer before the sign-function for features. A batch normalization layer ($BN$) consists of the moving average standard deviation $\sigma_{ma}$, a learnable scale $\gamma$ and a learnable shift factor $\beta$:
\begin{equation} \label{eq:bn}
    BN(Y) = \gamma \frac{Y-\mu}{\sqrt{\sigma_{ma}^2 + \epsilon}} + \beta
\end{equation}
where $Y$ is the output of convolution layer and input to a batch normalization layer.
\textbf{ReActNet (\citeyear{Liu2020ReActNet:Functions})} \cite{Liu2020ReActNet:Functions} used a channel-wise learnable bias (LB) before the sign-function for features.
Apart from a learnable bias, other works calculate the value of the bias based on other heuristics (DB). \textbf{ABC-Net (\citeyear{Lin2017TowardsNetwork})} \cite{Lin2017TowardsNetwork} calculates biases for both weights and features to minimize the quantization error in their ensemble approach. See the ensemble section for more information. \textbf{EqualBits (\citeyear{Li2021EqualWeights})} \cite{Li2021EqualWeights} uses the median of the weights as a bias value for the weights such that the information entropy is maximized. This implies the new weight distribution has an equal amount of weights below and above zero, and therefore after binarization, the amount of -1's and +1's is equal and entropy is maximized. \textbf{SD-BNN (\citeyear{Xue2021Self-DistributionNetworks})} \cite{Xue2021Self-DistributionNetworks} employs a series of linear layers and non-linearities to calculate biases for both weights and features. Note that this approach for the features belongs in the category of network topology changing.

\textbf{IR-Net (\citeyear{Qin2020ForwardNetworks})} \cite{Qin2020ForwardNetworks} normalizes the latent real-valued weights $W$ as follows (MSTD):
\begin{equation}
    W = \frac{W - \text{mean}(W)}{\sigma(W)}
\end{equation}
According to the authors, subtracting the mean\footnote{We think the authors have made a typing mistake in their paper and meant median instead of mean.} maximizes the information entropy of the weights and division by standard deviation makes the weights more dispersed. Hence, they claim that this normalization approach makes the real-valued weights steadily updated and the binary weights more stable during training.

\textbf{ReCU (\citeyear{Xu2021ReCU:Networks})} \cite{Xu2021ReCU:Networks} extends upon IR-Net's weight normalization by introducing a scaling factor $b$ for the standard deviation (MSTDB). In their experiments, a scaling factor equal of $b=\sqrt{2}$ works best. In addition, they normalize features as well by dividing them by the standard deviation (STD). Note that this does not affect the inference phase, because any multiplication being done before binarization does not affect the quantized output. Only the backward pass is affected: gradients will be scaled accordingly. Compared to a full batch normalization layer (equation \ref{eq:bn}), this approach does not include the moving average upon the standard deviation and is therefore different.

\subsubsection{Teacher-Student}
The Teacher-Student methodology aims to help train CNNs by having a more complex teacher network trying to transfer knowledge to a simple student network (the BNN). The knowledge transfer process can be done at distinct stages in the network. In all Teacher-Student approaches used in BNNs, the student binary network trains towards labels generated by the teacher network. No longer does it train towards the ground truth.
Occasionally, the teacher labels are mixed with the ground truth using a certain weight, as done in \cite{Bulat2019ImprovedRecognition}. On top of this Teacher-Student approach, it is also possible to use an additional loss term that is equal to the mean-squared error between the student's and teacher's building blocks.

\subsubsection{Regularization}
Regularization is a technique that introduces another loss term that should indirectly help train the network. \textbf{Regularized BNN (\citeyear{Darabi2018RegularizedTraining})} \cite{Darabi2018RegularizedTraining} proposed two regularization loss terms, based on the Manhatten distance ($R_1$) or the Euclidean distance ($R_2$), that try to push to a certain value $\pm \alpha$:
\begin{eqnarray}
    R_1 = \mid \alpha - \mid W \mid \mid \\
    R_2 = (\alpha - \mid W \mid)^2
\end{eqnarray}

\textbf{Information capacity in BNN (\citeyear{Ignatov2020ControllingNetwork})} \cite{Ignatov2020ControllingNetwork} introduced a regularization loss term $R_e$ that tries to maximize the binary weight ($W_B$) information entropy $H_f(W_B)$:
\begin{equation}
    R_e = \mid H_e - H_f(W_B) \mid
\end{equation}
where $H_e$ is the wanted information entropy and optimally equal to one.

\textbf{Increasing information entropy BNN (\citeyear{Zou2021IncreasingNetworks})} \cite{Zou2021IncreasingNetworks} extended this approach for the features, by employing a learnable bias before the feature binarizer that is constrained by a regularization loss term. This loss term tries to push the learnable bias towards the median of the features.

\subsubsection{Two-Stage Training}
The default in quantization-aware training is to train with both quantized features and weights simultaneously. In the two-stage training procedure, the first stage employs only binary features, whereas in the second stage both are binary. This training scheme is introduced by \textbf{Human pose estimation BNN (\citeyear{Bulat2019ImprovedRecognition})} \cite{Bulat2019ImprovedRecognition}. Note that in the second stage, the weight decay (L2 norm) is set to zero, which is crucial according to \cite{Liu2021HowOptimization}.

\subsubsection{Optimizer}
Within BNN works some use the SGD optimizer, whereas others use the ADAM optimizer, as can be seen in Table \ref{tab:overview}. Contrary to real-valued networks in which SGD supersedes ADAM in terms of accuracy, it is still under debate whether this is the case for BNNs. According to \cite{Liu2021HowOptimization}, ADAM's second momentum may play a crucial role. The SGD with momentum's update rule is:
\begin{eqnarray}
    m_t = m_{t-1} - \eta \frac{\delta L}{\delta W} \\
    W_{t+1} = W_{t} + m_t
\end{eqnarray}
where $t$ is the timestep, $W$ is the weight, $m$ is the first momentum, $\eta$ is the learning rate, and $l$ the network loss.
Similarly, ADAM's update rule is:
\begin{eqnarray}
    m_t = \beta_1 m_{t-1} + (1-\beta_1)\frac{\delta L}{\delta W} \\
    v_t = \beta_1 v_{t-1} + (1-\beta_2)(\frac{\delta L}{\delta W})^2 \\
    \hat{m}_t = \frac{m_t}{1-\beta_1^t} \\
    \hat{v}_t = \frac{v_t}{1-\beta_2^t} \\
    W_t = W_{t-1} - \eta \frac{\hat{m}_t}{\sqrt{\hat{v}_t+\epsilon}} \\
\end{eqnarray}
where $v$ is the second momentum, $\beta_1$ and $\beta_2$ are the first and second momentum coefficient, and $\epsilon$ a small value to prevent division by zero. 

Next to the default optimizers, \cite{Helwegen2019LatentOptimization,Suarez-Ramirez2021ANetworks} have developed new optimizers that are dedicated to BNNs, which are respectively called Bop and Bop2ndOrder.

\subsection{Network topology changing}
BNN accuracy repairs can also be obtained by changing the network topology. The repair method branch can be split up into the following categories: scaling factor, ensemble, activation function, double residual, and squeeze-and-excitation. Each repair method within these categories is outlined in the sections below. Note that these repair methods enlarge the network topology and therefore will add (real- and/or binary-valued) operations to the original network.
\subsubsection{Scaling factor}
\textbf{XNOR-Net (\citeyear{Rastegari2016XNOR-Net:Networks})} \cite{Rastegari2016XNOR-Net:Networks} wanted to minimize the quantization error between the real- and binary-valued weights using a channel-wise scaling factor $\alpha$:
\begin{equation}
    \min\limits_{\alpha} \mid W_R - \alpha W_B \mid^2
\end{equation}
Their mathematical derivation led to the result that $\alpha$ should be equal to the average of the absolute weights for each channel.

\textbf{XNOR-Net++ (\citeyear{Bulat2019XNOR-Net++:Networks})} \cite{Bulat2019XNOR-Net++:Networks} made the scaling factor a learnable parameter (LF). Moreover, it explored the dimensionality of the scaling factor.
\textbf{EWGS (\citeyear{Lee2021NetworkScaling})} \cite{Lee2021NetworkScaling} extended the learnable parameter with a dedicated intialization. It initialized the scaling factor with the ratio between the average absolute real output and average absolute quantized output.

\subsubsection{Ensemble}
\begin{figure}
  \centering
  \includegraphics[width=\textwidth]{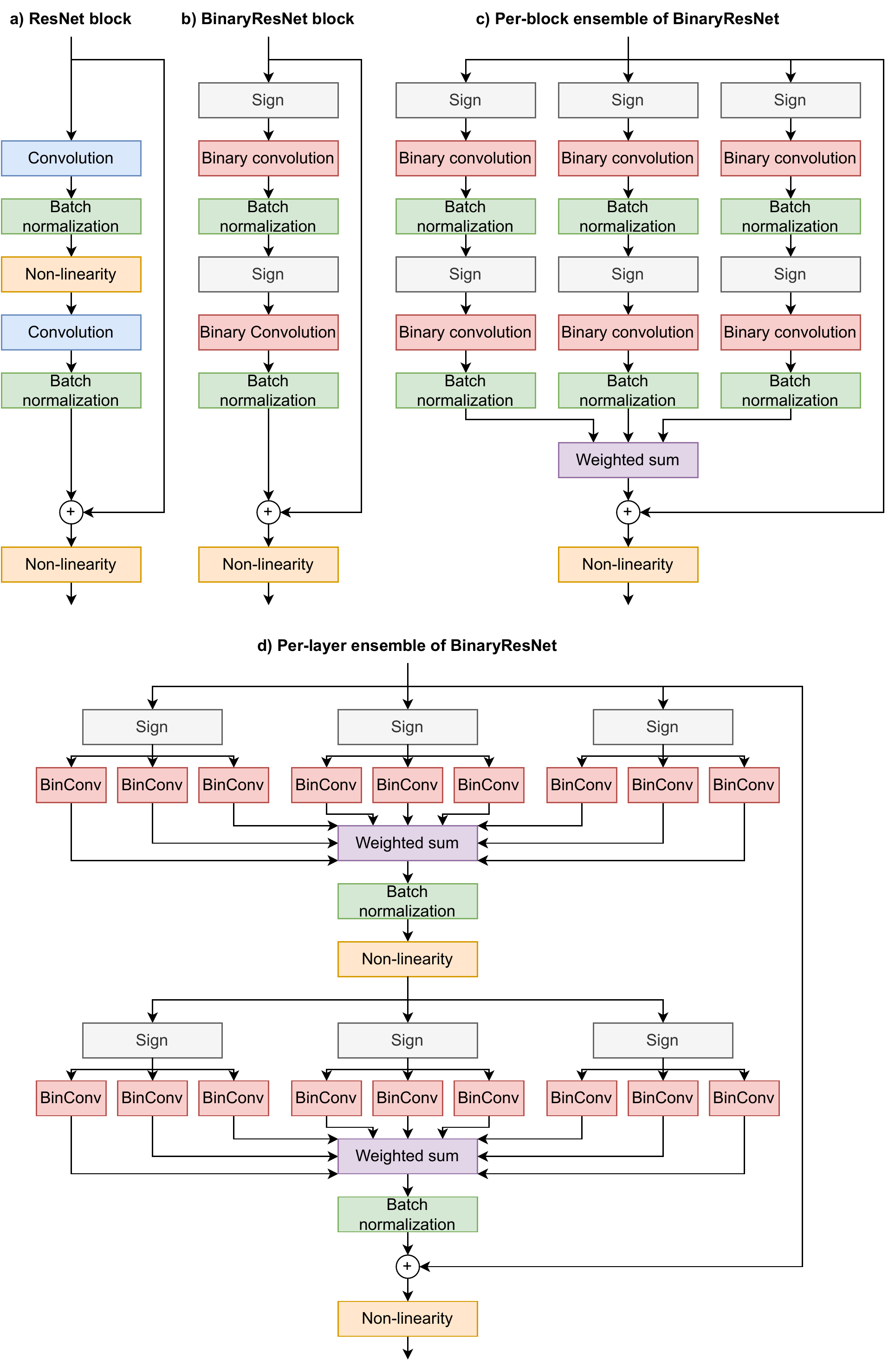}
  \caption{Ensemble approach applied to ResNet building blocks: (a) real-valued ResNet, (b) binary ResNet, (c) per-layer ensemble binary ResNet, and (d) per-block ensemble binary ResNet. Note that binary feature map data only exists between the sign activation and binary convolution.}
  \label{fig:topology_changing_ensemble}
\end{figure}
Ensemble methods use multiple instantiations of (a certain part of) the network to obtain better performance than could be obtained from any individual part. We can distinguish three distinct ensembles: per-layer, per-block, and per-network. In Fig. \ref{fig:topology_changing_ensemble} examples of per-layer and per-block ensembles are illustrated based on the ResNet architecture.

\textbf{ABCNet (\citeyear{Lin2017TowardsNetwork})} \cite{Lin2017TowardsNetwork} is based on the per-layer ensemble, which converts a single convolution $N$ with real-valued input and weights to a binary ensemble of $N^2$ binary convolutions. The binary convolutions are weighted and summed at the end to get a similar feature size as the original.

\textbf{GroupNet (\citeyear{Zhuang2019StructuredSegmentation})} \cite{Zhuang2019StructuredSegmentation} is based on the per-block ensemble. For every building block of a certain network architecture, the ensemble is formed. Similar to ABCNet \cite{Lin2017TowardsNetwork}, a weighted sum is used to get to the original feature size. Compared to the per-layer ensemble this ensemble expands the network by a factor $N$ rather than $N^2$.

\textbf{BENN (\citeyear{Zhu2019BinaryBit})} \cite{Zhu2019BinaryBit} is based on the per-network ensemble. Similar to the other approaches a weighted sum is used to combine individual instantiations of the network. Compared to the per-group approach this only saves in the number of weighted sums to do.

\subsubsection{Activation function}
\begin{figure}
  \centering
  \includegraphics[width=0.8\textwidth]{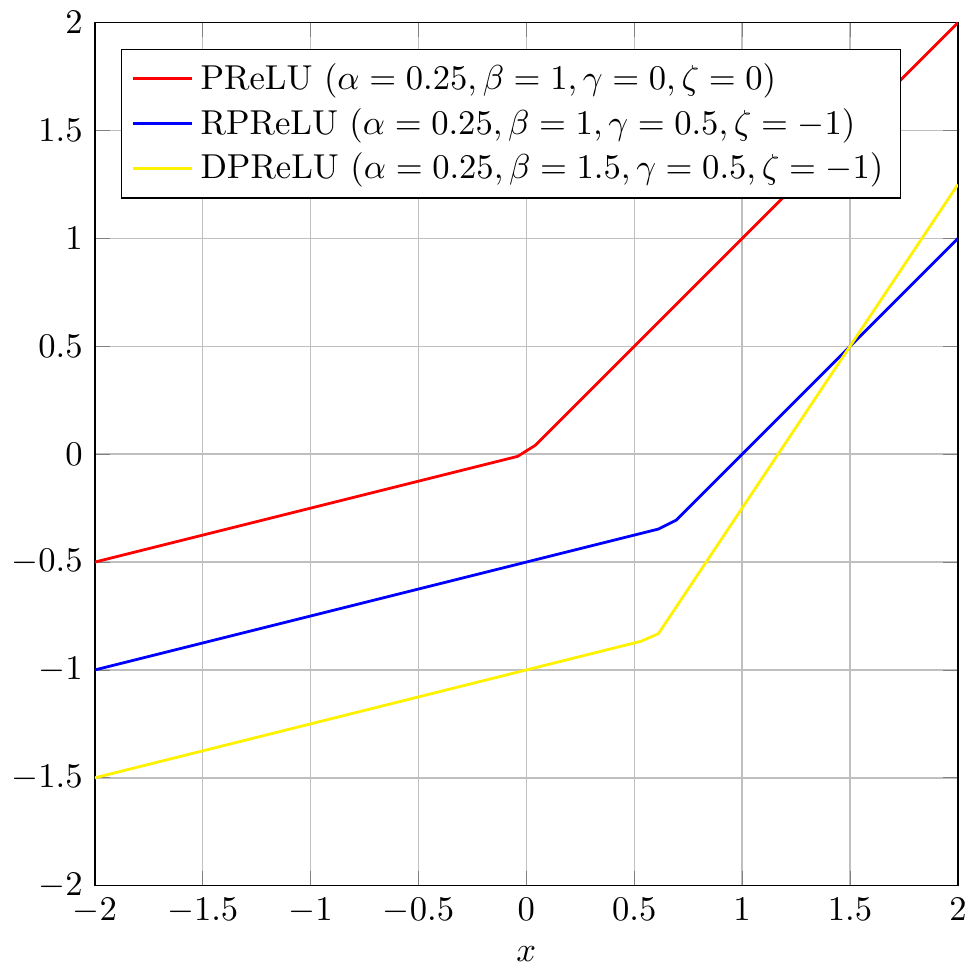}
  \caption{Three PReLU variants illustrated: PReLU, RPReLU, and DPReLU. The general equation is shown in equation \ref{eq:prelus}.}
  \label{fig:prelus}
\end{figure}

Next to the binarizer that can be regarded as the non-linearity in the network, several BNN works have suggested to add an additional activation function to improve the accuracy: ReLU, PReLU, or a variant upon PReLU. As can be seen in Table \ref{tab:overview}, the PReLU is quite commonly used. \textbf{ReActNet (\citeyear{Liu2020ReActNet:Functions})} \cite{Liu2020ReActNet:Functions} introduced the RPReLU, which is a regular PReLU preceded and proceeded by a channel-wise learnable bias. \textbf{PokeBNN (\citeyear{Zhang2021PokeBNN:Accuracy})} \cite{Zhang2021PokeBNN:Accuracy} extended the RPReLU by allowing the positive slope to be learnable as well. An example of all PReLU variants is shown in Fig. \ref{fig:prelus}. A general equation these variants is:
\begin{equation}\label{eq:prelus}
    max(\alpha X, \beta(X-\gamma)) + \zeta
\end{equation}
where $X$ is the input, $\alpha$ and $\beta$ are respectively the negative and positive slope, and $\gamma$ and $\zeta$ are respectively the bias before and after PReLU.

\subsubsection{Double residual}
\begin{figure}
  \centering
  \includegraphics[width=0.9\textwidth]{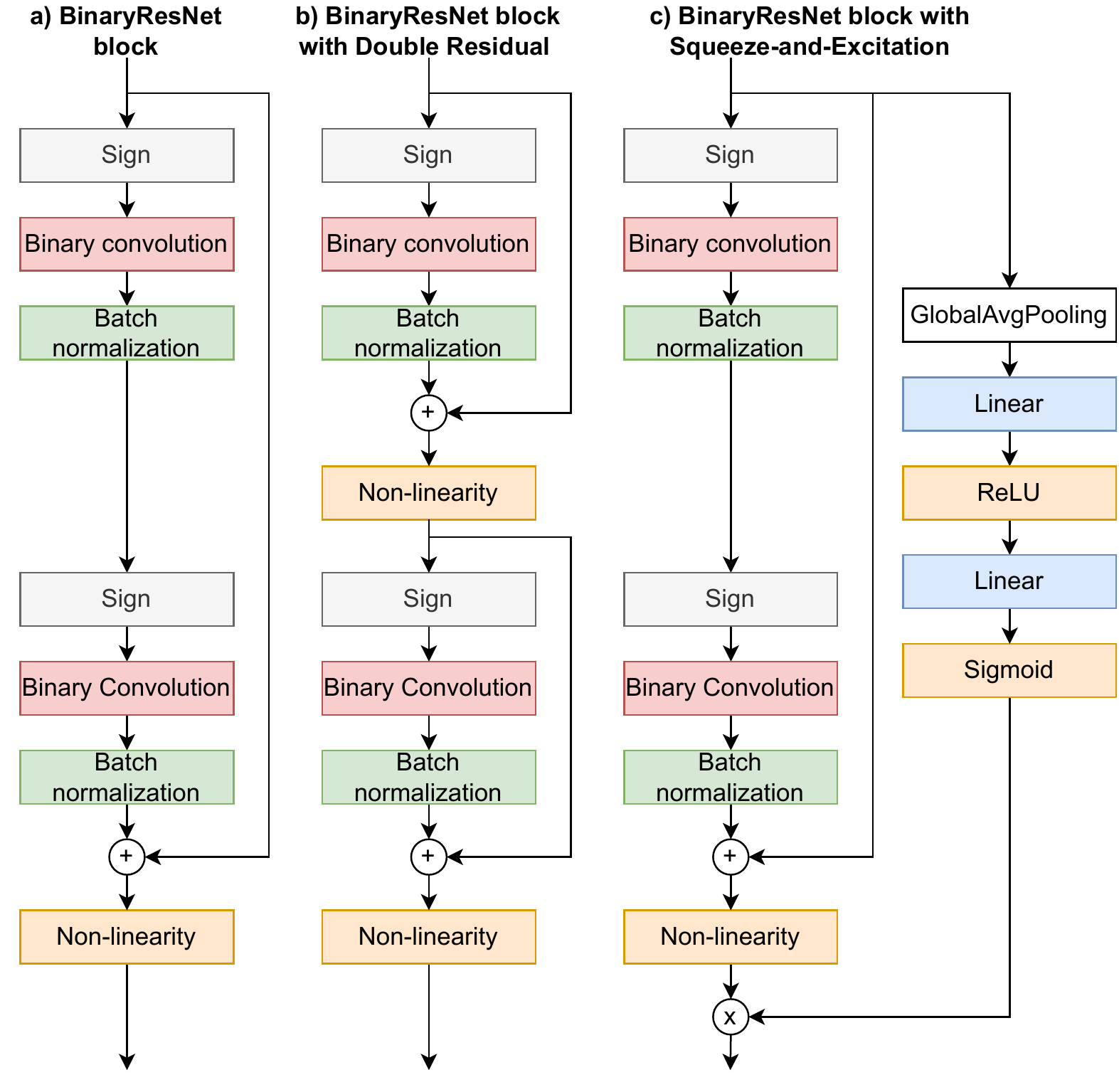}
  \caption{Two network topology changing repairs shown on top of binary ResNet's building block (a): double residual repair (b) and Squeeze-and-Excitation repair (c).}
  \label{fig:topology_changing_2r_se}
\end{figure}
To keep the flow of information rich in the network, BNNs use a residual connection after every convolution (2R), instead of after two convolutions as in the usual ResNet architecture. This idea was proposed by \textbf{Bi-RealNet (\citeyear{Liu2018Bi-RealAlgorithm})} \cite{Liu2018Bi-RealAlgorithm} and used by many later works. The default ResNet block and the new topology are illustrated in Fig. \ref{fig:topology_changing_2r_se}. Note that while this does add real-valued additions, it reduces the lifetime of high-precision features, which potentially may increase the energy efficiency when mapped to an embedded processing system. 

\subsubsection{Squeeze-and-Excitation}
The concept of Squeeze-and-Excitation (SE) is to assign each output feature map channel a different weight based on its importance. These channel-wise weights are based on the input feature. An example of a Squeeze-and-Excitation block on top of the ResNet architecture is shown in Fig. \ref{fig:topology_changing_2r_se}. Note that the position of the residual addition and SE multiplication differs per BNN work. In Fig.\ref{fig:topology_changing_2r_se}, the method of \textbf{PokeBNN (\citeyear{Zhang2021PokeBNN:Accuracy})} \cite{Zhang2021PokeBNN:Accuracy} is shown. In \textbf{Real-to-Binary (\citeyear{Martinez2020TrainingConvolutions})} \cite{Martinez2020TrainingConvolutions}, for example, the order is switched. Note that Squeeze-and-Excitation adds many real-valued operations (approximately equal to twice the square of the feature map channels), both multiplications and additions. 

\section{Empirical review of accuracy repair methods} \label{sec:review}
The holy grail for deploying CNNs on resource-constrained embedded systems may be BNNs since they save a lot on memory size and throughput and simplify computation by binarizing both features and weights. Binarization inevitably causes severe accuracy loss and BNN's intrinsic discontinuity brings difficulty to its training. Fortunately, many training techniques and network topology changes have been proposed that aim to reduce the accuracy loss, as shown in Section \ref{sec:classification} and \ref{sec:overview}. Problem, however, is that BNN literature does not agree on what improvements should be used to get high-accuracy BNNs.
As can be seen in our overview table in the previous section, Table \ref{tab:overview}, each BNN work has its own series of accuracy repair techniques to become more accurate on the image classification task. Therefore, we will evaluate the benefits of each repair method in isolation by providing empirical proof over several datasets and model architectures. 

Our approach to evaluating each repair method in isolation consists of three steps:
\begin{enumerate}
    \item Establish the design space of all individual repair methods that we will evaluate in this chapter.
    \item Finding a good baseline BNN and the hyperparameters to use
    \item Explore the design space using the found hyperparameters of the previous step
\end{enumerate}

As can be seen in Table \ref{tab:overview} the majority of BNN works report their achieved accuracy on ResNet-18 architecture with the ImageNet dataset. Doing all of our experiments on ImageNet is not possible, since the design space combined with the experimental setup would require too many GPU-hours. Hence, for each experiment that we do, we train the model on two smaller benchmarks: ResNet-20 with the CIFAR10 dataset (R20C10) and ResNet-18 with the CIFAR100 dataset (R18C100). Moreover, we train each experiment five times with five different seeds. This way, we can make claims about best, mean, and deviation in test accuracy with respect to a certain repair method. A successful repair method should not only get high accuracy, but its deviation should be low as well to avoid the need to train many times while hoping to be lucky.

In Section \ref{sec:review:ds} we establish the design space that is used in the review. In Section \ref{sec:review:baseline} a baseline is established for the benchmarks, while in Section \ref{sec:review:dse} the results of our evaluation are shown.

\subsection{Establishing the design space}\label{sec:review:ds}
In order to evaluate each individual repair method, we have organized and classified them in Table \ref{tab:classification} in Section \ref{sec:classification}. However, for practical reasons we have limited the design space to further reduce the amount of required GPU-hours\footnote{We have spent over 4000 GPU-hours to get the results of our review.}. We exclude the following categories in this work for the following reasons:
\begin{itemize}
    \item \textbf{Normalization}: from this category we only exclude the dynamic bias (DB) method, because this method has three different implementations and mostly requires many real-valued operations, which would be costly on embedded systems.
    \item \textbf{Teacher-Student}: Teacher-Student knowledge transfer has been shown to be effective. However, the approaches taken are not specific to BNN training.
    \item \textbf{Regularization}: additional loss terms are rarely used.
    \item \textbf{Optimizer}: evaluating multiple optimizers is costly in the number of trainings to do since it linearly increases the number of experiments to conduct. Thus, in this work, we chose to exclude all custom BNN optimizers and only focus on the optimizers that most works are using: ADAM and SGD. Furthermore, to reduce the design space, even more, we chose one optimizer for the design space exploration based on the results in the "finding a good baseline" subsection.
    \item \textbf{Ensemble}: this category of repairs will introduce many additional binary operations and introduce real-valued weighted sums as well. Since the focus of BNNs is on resource-constrained embedded systems, we feel this category should not be explored.
    \item \textbf{Squeeze-and-Excitation}: few BNN works use this repair method and those that do differ in the implementation. Moreover, this block introduces only real-valued operations, which would be costly on embedded systems.
\end{itemize}

\begin{table}[]
\caption{Design space used in our empirical review}
\label{tab:designspace}
\begin{tabular}{@{}llll@{}}
\toprule
 & \textbf{Property} & \textbf{Abbreviation} & \textbf{Description} \\ \midrule
\multirow{22}{*}{\textbf{\rotatebox[origin=c]{90}{Training technique}}} & \multirow{8}{*}{Binarizer (STE)} & LC\_$|X|$ & Linear clipped at $|X|$ \\
 &  & LC\_A & Linear with adaptive clipping \\
 &  & PN & Polynomial \\
 &  & GPN & Gradual Polynomial \\
 &  & T & Gradual tanh-based sign \\
 &  & EDE & T with magnitude scaling \\
 &  & SS & SwishSign \\
 &  & EWGS & Element-wise Gradient Scaling \\\cmidrule(l){2-4} 
 & \multirow{6}{*}{Normalization} & LB & Learnable bias \\
 &  & STD & Division by standard deviation \\
 &  & MSTD & Zero-mean and division ny standard deviation \\
 &  & MSTDB & MSTD divided by $b$ \\
 &  & BN & Batch normalization \\\cmidrule(l){2-4} 
 & \multirow{2}{*}{Optimizer} & SGD & SGD optimizer \\
 &  & ADAM & ADAM optimizer \\\toprule
\multirow{13}{*}{\textbf{\rotatebox[origin=c]{90}{Topology changing}}} & \multirow{3}{*}{Scaling factor} & AM & Absolute mean of the Weights \\
 &  & LF & Learnable factor \\
 &  & LFI & Learnable factor with initialization \\\cmidrule(l){2-4} 
 & \multirow{5}{*}{Non-linearity} & I\&H & Identity and htanh \\
 &  & ReLU & ReLU \\
 &  & PReLU & PReLU \\
 &  & RPReLU & RPReLU \\
 &  & DPReLU & DPReLU \\\cmidrule(l){2-4} 
 & Double residual & 2R\_Y/N & Residual per convolution Yes / No \\\midrule
\end{tabular}
\end{table}

\begin{figure}
  \centering
  \includegraphics[width=\textwidth]{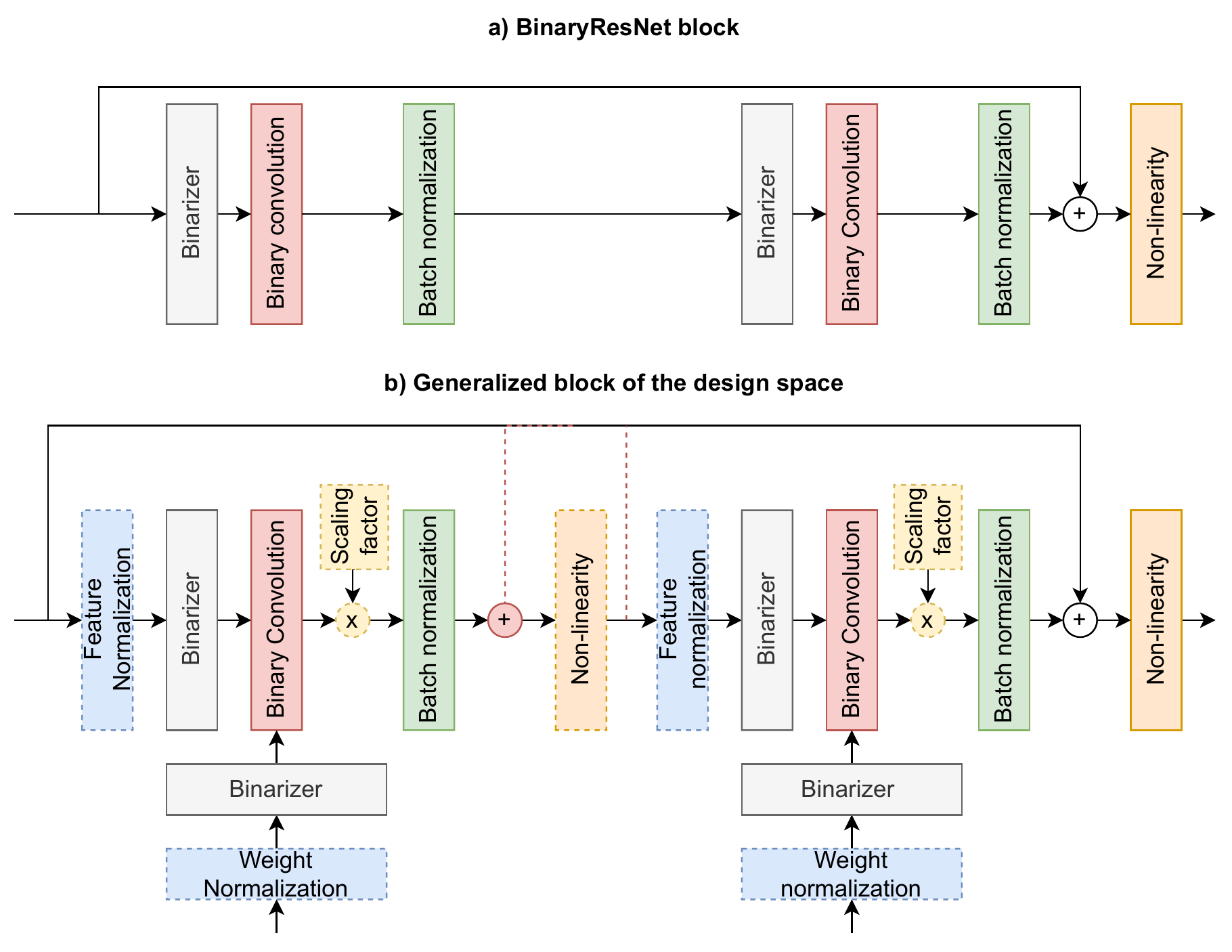}
  \caption{Comparison of (a) binary ResNet's building block and (b) our generalized building block. The generalized building block (b) is constructed by mapping the design space, listed in Table \ref{tab:designspace}, on top of (a). Dashed lines indiciate the new operations.}
  \label{fig:generalized_building_block}
\end{figure}

As such the design space that will be evaluated is shown in Table \ref{tab:designspace}. Note that the binarizer and normalization categories can be applied to both features and weights and therefore increase the design space. To simplify understanding we visualized the design space in Fig. \ref{fig:generalized_building_block} in which the generalized building block is shown. This building block is an extension of the building block used in BinaryResNet \cite{Courbariaux2016Binarized-1}, the very first BNN work published. The only categories that cannot be shown are two-stage training and optimizer since they do not introduce any new operations within the model.

\subsection{Finding a good baseline BNN}\label{sec:review:baseline}
Since there are no test accuracy numbers available for our benchmarks (R20C10 and R18C100), a good baseline has to be established. Moreover, good hyperparameters should be found and used for the actual design space exploration. In this subsection, we want to explore how many epochs a BNN should train for, what data augmentation should be used, and which optimizer is best.

For this series of experiments, we fix the learning rate scheduler as cosine decay with a warm-up of two epochs. We initialize the networks using Kaiming normal initialization \cite{He2015DelvingClassification}, use a weight decay of $10^{-4}$, and use a batch size of 128. The data augmentation is either random horizontal flips, random crops, and normalization (default), or includes AutoAugmentation \cite{Cubuk2018AutoAugment:Data} after the random crops (AutoAug). Regarding the optimizers, SGD and ADAM, we use the following (default) hyperparameters: a learning rate of 0.1 and a first momentum coefficient of 0.9 for SGD, whereas for ADAM we use a learning rate of $10^{-3}$, first momentum coefficient of 0.9, and second momentum coefficient of 0.999.

\begin{figure}
  \centering
  \includegraphics[width=\textwidth]{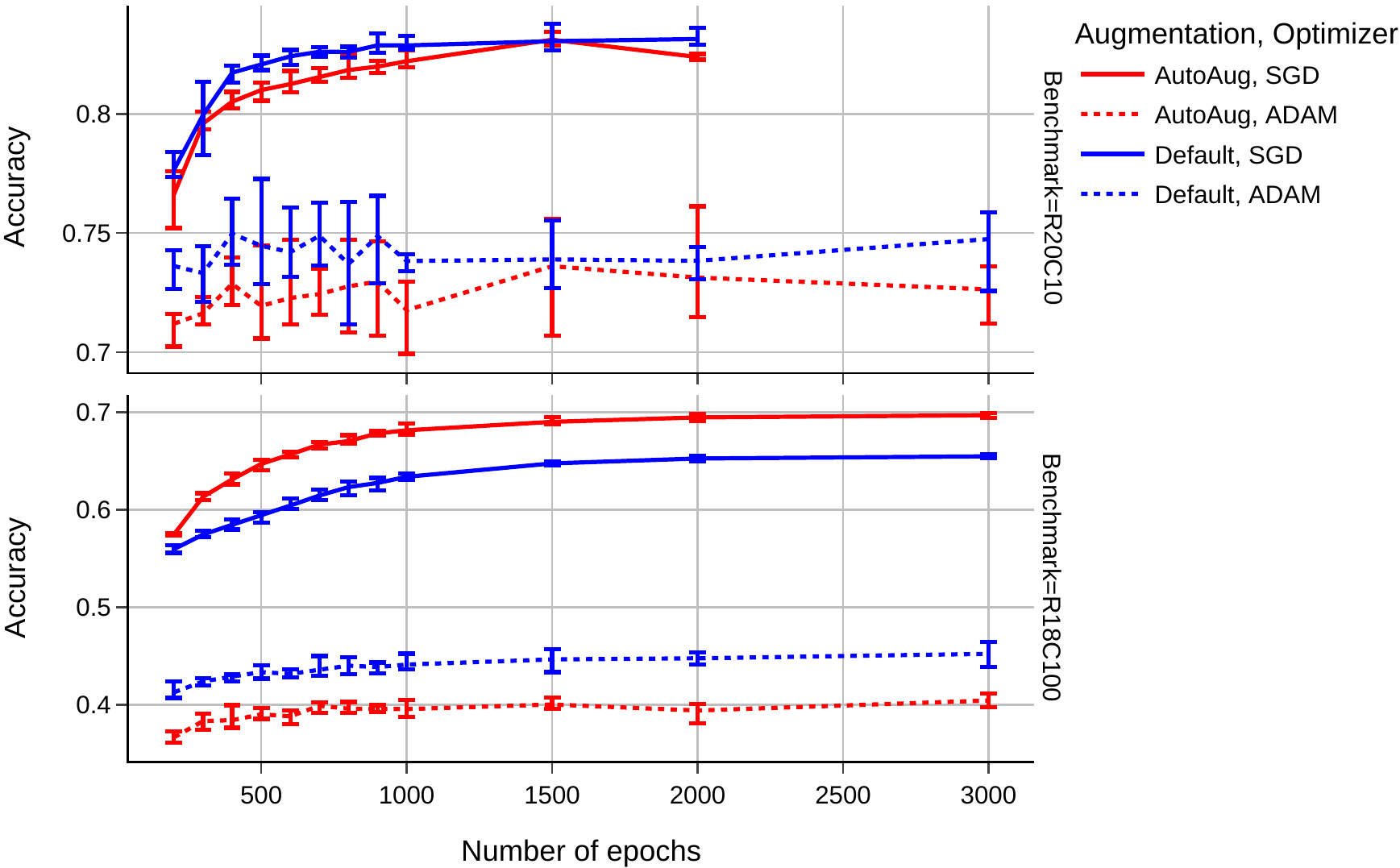}
  \caption{Baseline accuracy on binary versions of R20C10 and R18C100 with varying amount of epochs. Dashed lines show ADAM optimizer, whereas a normal line is SGD optimizer. Red line is usage of AutoAug data augmentation, while blue line is default data augmentation.}
  \label{fig:baseline}
\end{figure}
Fig. \ref{fig:baseline} illustrates all possible combinations between epochs, data augmentation and optimizer. It can be seen that there is a difference between the R20C10 and R18C100 experiments. Regarding the optimizer choice, we see that SGD outperforms ADAM consistently. Regarding data augmentation, it is best to use default data augmentation for R20C10 and AutoAug for R18C100. Lastly, at some point training for more epochs does not increase the accuracy by a lot anymore. Thus, to keep training time limited we take the number of epochs within the 1\%-point of the maximum achievable result, which implies 500 epochs for R20C10 and 1000 epochs for R18C100.

To put the baseline results into perspective, Table \ref{tab:accuracies} reports the accuracies of the benchmarks as reported in the literature. On R20C10 we achieve a maximum of 83.3\%, which seems to be in the lower end, but still reasonable as it is a baseline and does not have any repairs. On the R18C100 benchmark, it is different. We achieve a maximum of 69.8\% which already beats some works that include repairs, but still, there seems room for repairs. Overall, it seems that by the usage of repairs methods the accuracy can be raised by 6\%-points.

\begin{table}[]
\caption{Accuracy of ResNet-20\&CIFAR10 (R20C10) and ResNet-18\&CIFAR100 (R18C100) in literature and compared to our baseline.}
\label{tab:accuracies}
\begin{tabular}{lllll}
\hline
\textbf{Work} & \textbf{\begin{tabular}[c]{@{}l@{}}R20C10 \\ Accuracy\end{tabular}} & \textbf{} & \textbf{Work} & \textbf{\begin{tabular}[c]{@{}l@{}}R18C100 \\ Accuracy\end{tabular}} \\ \hline
Full-precision \cite{He2015DeepRecognition} (our training) & 92.71\% & \hspace{2.5em} & Full-precision \cite{He2015DeepRecognition} (our training) & 78.20\% \\
Our baseline & 83.3\% &  & Our baseline & 69.8\% \\
CI-CBNN \cite{Wang2019LearningNetworks} & 91.10\% &  & Circulant BNN \cite{Liu2019CirculantPropagation} & 69.97\% \\
IR-Net \cite{Qin2020ForwardNetworks} & 86.5\% &  & Real-to-Binary \cite{Martinez2020TrainingConvolutions} & 76.2\% \\
BBG \cite{Shen2020BalancedResidual} & 85.34\% &  & ProxyBNN \cite{He2020ProxyBNN:Matrices} & 67.17\% \\
RBNN \cite{Lin2020RotatedNetwork} & 87.8\% &  & Information capacity BNN \cite{Ignatov2020ControllingNetwork} & 73.48\% \\
Noisy Supervision \cite{Han2020TrainingSupervision} & 85.78\% &  & ReCU \cite{Xu2021ReCU:Networks} & 69.1\% \\
ReCU \cite{Xu2021ReCU:Networks} & 87.4\% &  & BNN-BN \cite{Chen2021BNNNormalization} & 68.34\% \\
Sub-bit BNN \cite{Wang2021Sub-bitNetworks} & 83.9\% &  & Equal Bits \cite{Li2021EqualWeights} & 71.60\% \\
SD-BNN \cite{Xue2021Self-DistributionNetworks} & 86.9\% &  &  &  \\
BNN fully latent weights \cite{Xu2021ImprovingWeights} & 88.6\% &  &  &  \\ \hline
\end{tabular}
\end{table}

\subsection{Design space exploration} \label{sec:review:dse}
Based on the baseline hyperparameter study in the previous section we will use SGD as the optimizer. R20C10 experiments will use default data augmentation and train for 500 epochs, whereas R18C100 experiments will use AutoAug data augmentation and train for 1000 epochs.

\subsubsection{Binarizer (STE)}
\begin{figure}
  \centering
  \includegraphics[width=\textwidth]{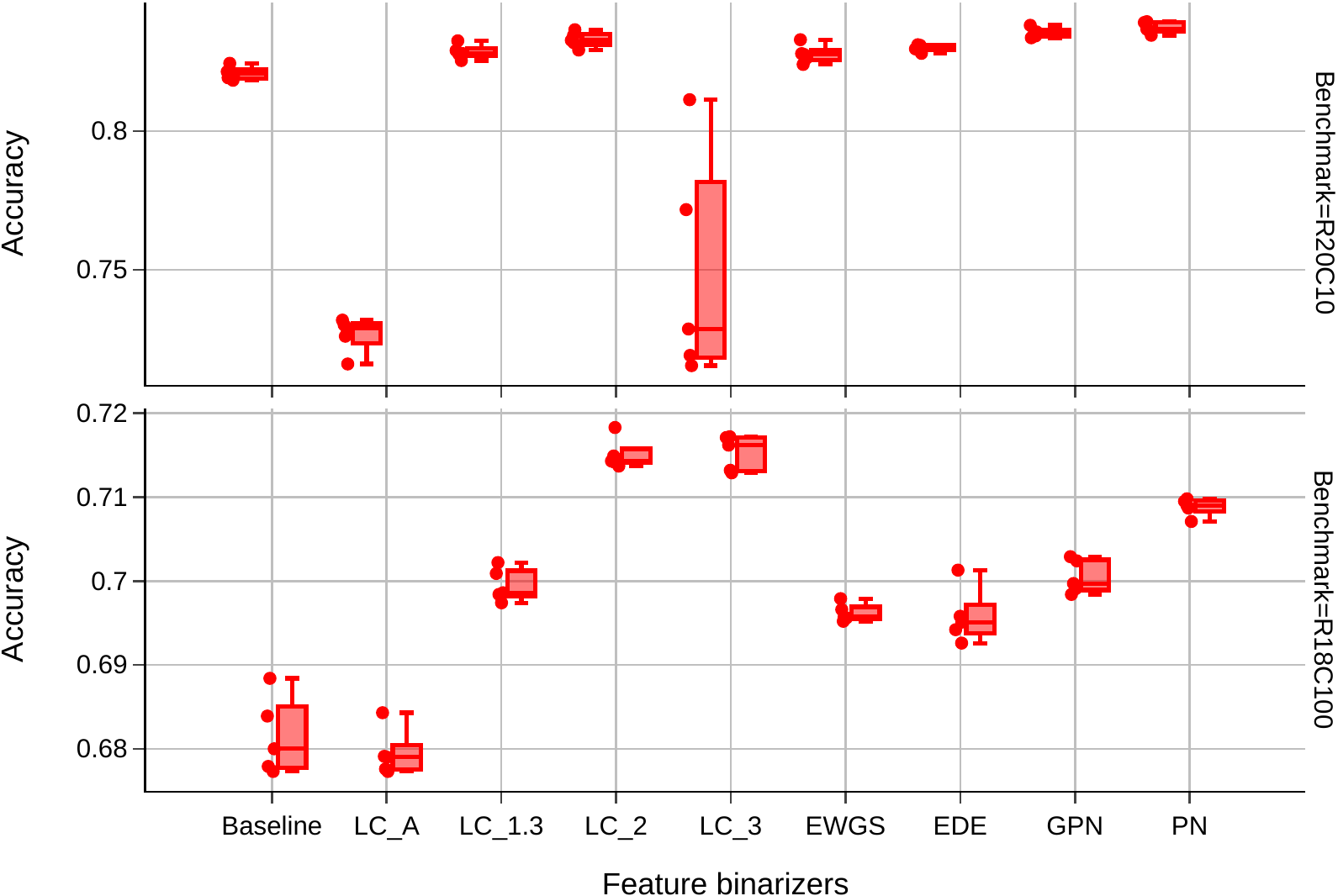}
  \caption{Baseline BNN compared to different feature binarizers. SS and T binarizers have been excluded due to poor performance. Dots indicate measurement points and the box-and-whiskers shows the minimum, median and maximum of the five measurements per experiment.}
  \label{fig:dse_fste}
\end{figure}
Fig. \ref{fig:dse_fste} illustrates the different feature binarizers compared to the baseline. The SS and T binarizer have been excluded in this figure, because of poor performance (more than 20\%-point below baseline). These two binarizers share one commonality: they produce large gradient magnitudes for feature values close to zero. For the T binarizer, this occurs when $\lambda$ has grown too large during training. Hence, we think the $\lambda$ values reported in \cite{Bulat2019ImprovedRecognition} are not correct, or other crucial information about the binarizer is missing.

The extension of the T binarizer, EDE, scales the magnitude using $\lambda$ as well and gives a slight improvement over the baseline.

The LC series of binarizers show mixed results. The adaptive LC\_A binarizer does not help in the R18C100 benchmark but reduces the accuracy in the R20C10 benchmark by almost 10\%-point. Enlarging the clipping interval seems to help a lot, up to a 3\%-point increase. The LC\_2 binarizer achieves the highest accuracy in both benchmarks. Further enlarging the interval to LC\_3 results in a large deviation in accuracy, especially for R20C10.

Compared to the baseline the EWGS binarizer gives a slight improvement, but not as much as others. Most notable is the decrease in deviation in the R18C100 benchmark.

The PN binarizer, with its triangular derivative shape, is good in both benchmarks. In R20C10 its the highest scoring, while in the R18C100 benchmark it is third-best after LC\_2 and LC\_3. Its gradual counterpart, GPN, does not improve upon PN, which can clearly be seen in the R18C100 graph.

Overall it seems that binarizers that are fixed during training have the best performance. The LC (with clipping interval larger than 1) and PN binarizers for features should be used when training BNNs for embedded systems.

\begin{figure}
  \centering
  \includegraphics[width=\textwidth]{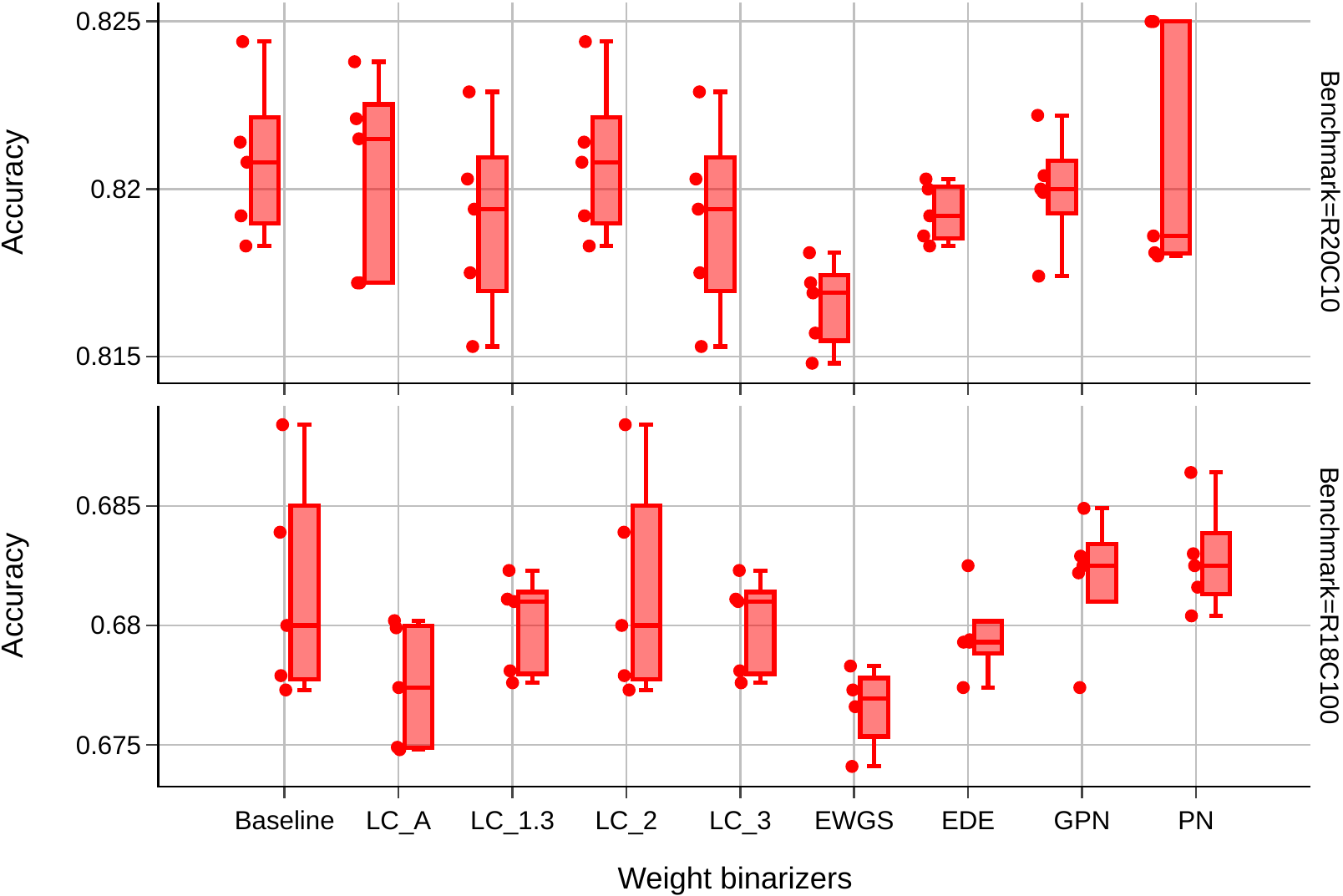}
  \caption{Different weight binarizers compared to the baseline on two benchmarks. SS and T binarizers have been excluded due to poor performance.}
  \label{fig:dse_wste}
\end{figure}
Next to the binarizers for features, the same binarizers have been applied to the weight for which the results are shown in Fig. \ref{fig:dse_wste}. Similar to the feature binarizers, the SS and T binarizers underperform. Contrary to the features, however, there are no major improvements in test accuracy when applying a different binarizer for the weights. This could be logical because during backpropagation the gradient error gets accumulated when binarizing features.

\subsubsection{Normalization}
\begin{figure}
  \centering
  \includegraphics[width=\textwidth]{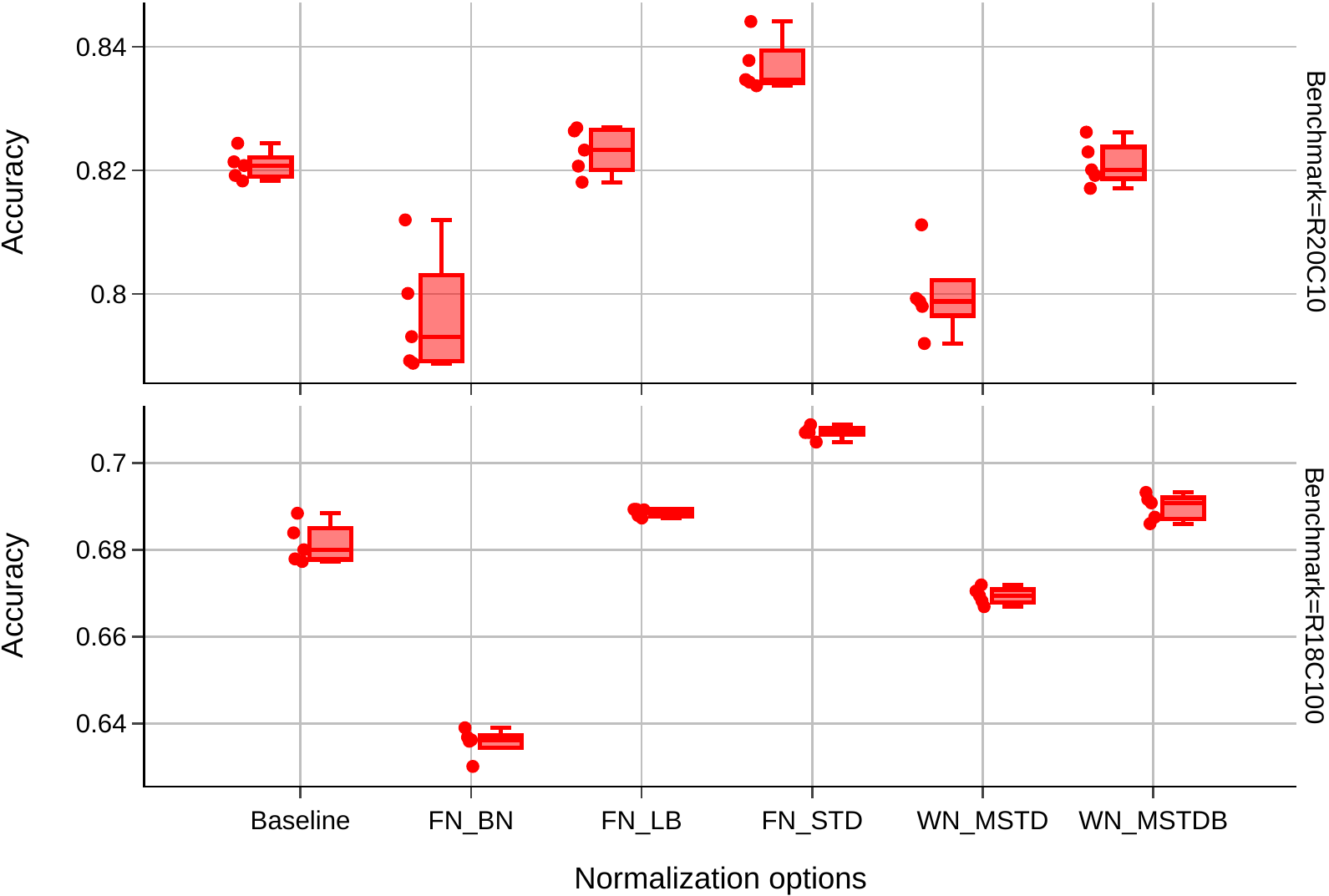}
  \caption{Various feature and weight normalization options compared to the baseline on two benchmarks.}
  \label{fig:dse_normalization}
\end{figure}

Fig. \ref{fig:dse_normalization} illustrates the different normalization methods for both features and weights. Feature normalization is denoted by the prefix FN, and weight normalization using the prefix WN. 

Using LB as feature normalization does not result in an improvement in accuracy, and regarding deviation, it has mixed results. In R18C100 the deviation decreases, whereas in R20C10 it increases.

Contrary to LB normalization, STD feature normalization results in a roughly 2\%-point increase in accuracy in both benchmarks, which makes it a very interesting parameter. STD normalization probably helps the optimization of BNNs by solving the gradient vanishing or exploding problem. Rescaling gradients by the standard deviation, therefore, ensures proper learning capabilities.

An interesting observation is that using BN as normalization is destructive in performance, even though it is a combination of STD and LB. However, there is one crucial difference, and that is the moving average on top of the standard deviation (as shown in equation \ref{eq:bn}).

In the weight normalization part, we have two very similar options: MSTD and MSTDB. Fig. \ref{fig:dse_normalization} shows that MSTD reduces performance in both benchmarks, whereas MSTDB only has a negligible improvement in the R18C100 benchmark. Thus, controlling the standard deviation factor $b$ seems to help but does not improve upon the baseline.

\subsubsection{Scaling factor}
\begin{figure}
  \centering
  \includegraphics[width=\textwidth]{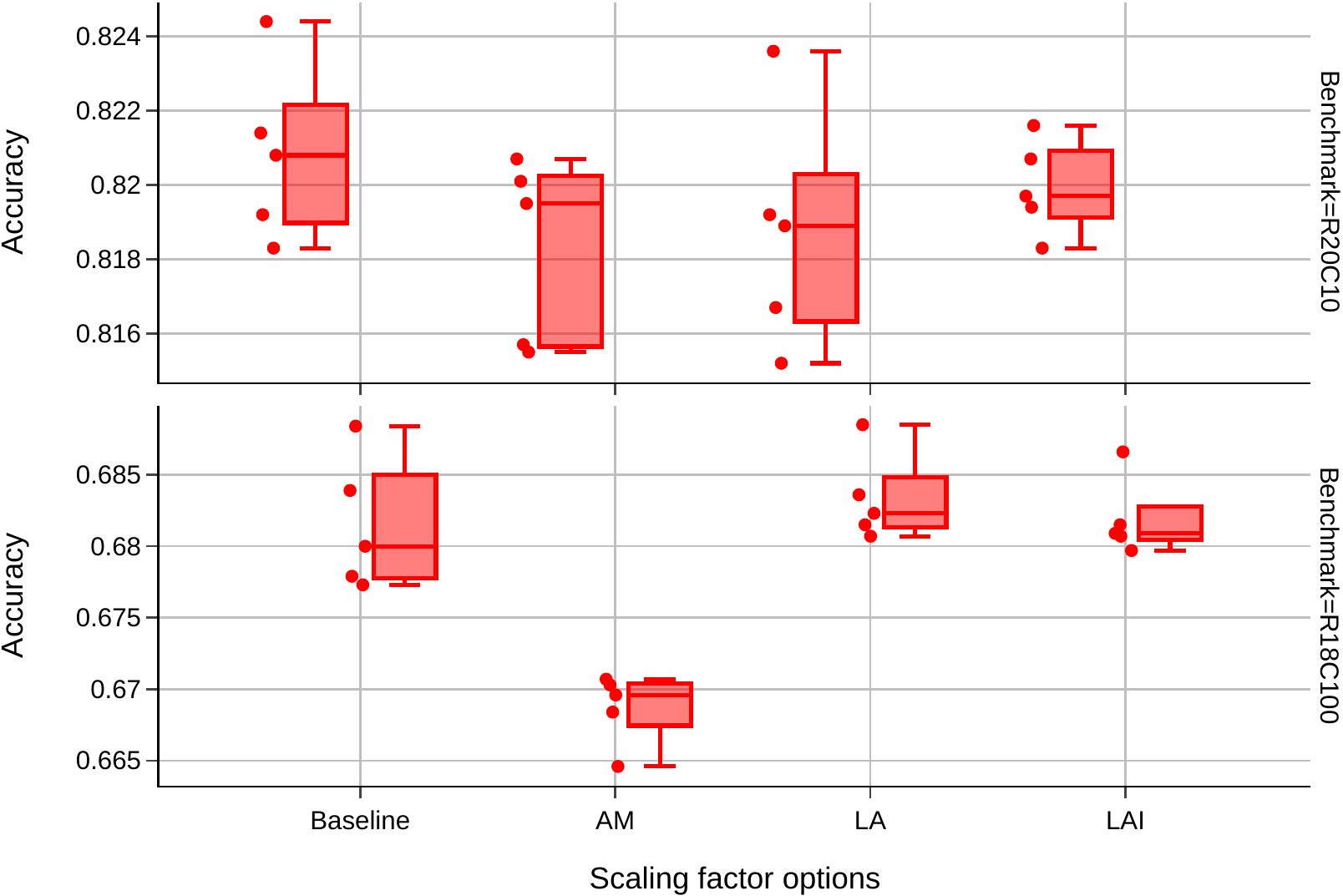}
  \caption{On two benchmarks diverse scaling factor methods are shown and compared to the baseline.}
  \label{fig:dse_sf}
\end{figure}

Fig. \ref{fig:dse_sf} illustrates the various scaling factor methods. Note the range of the y-axis on both benchmarks, it is rather small, which already indicates the negligible effect of the scaling factor category. This is expected since any scaling factor would be absorbed by the batch normalization layer that proceeds the scaling factor. Differences in accuracy between the baseline and the scaling factors are probably due to the non-deterministic behavior of floating-point calculations. Thus, this repair method category is not needed.

\subsubsection{Two-stage training, activation function and double residual}
\begin{figure}
  \centering
  \includegraphics[width=\textwidth]{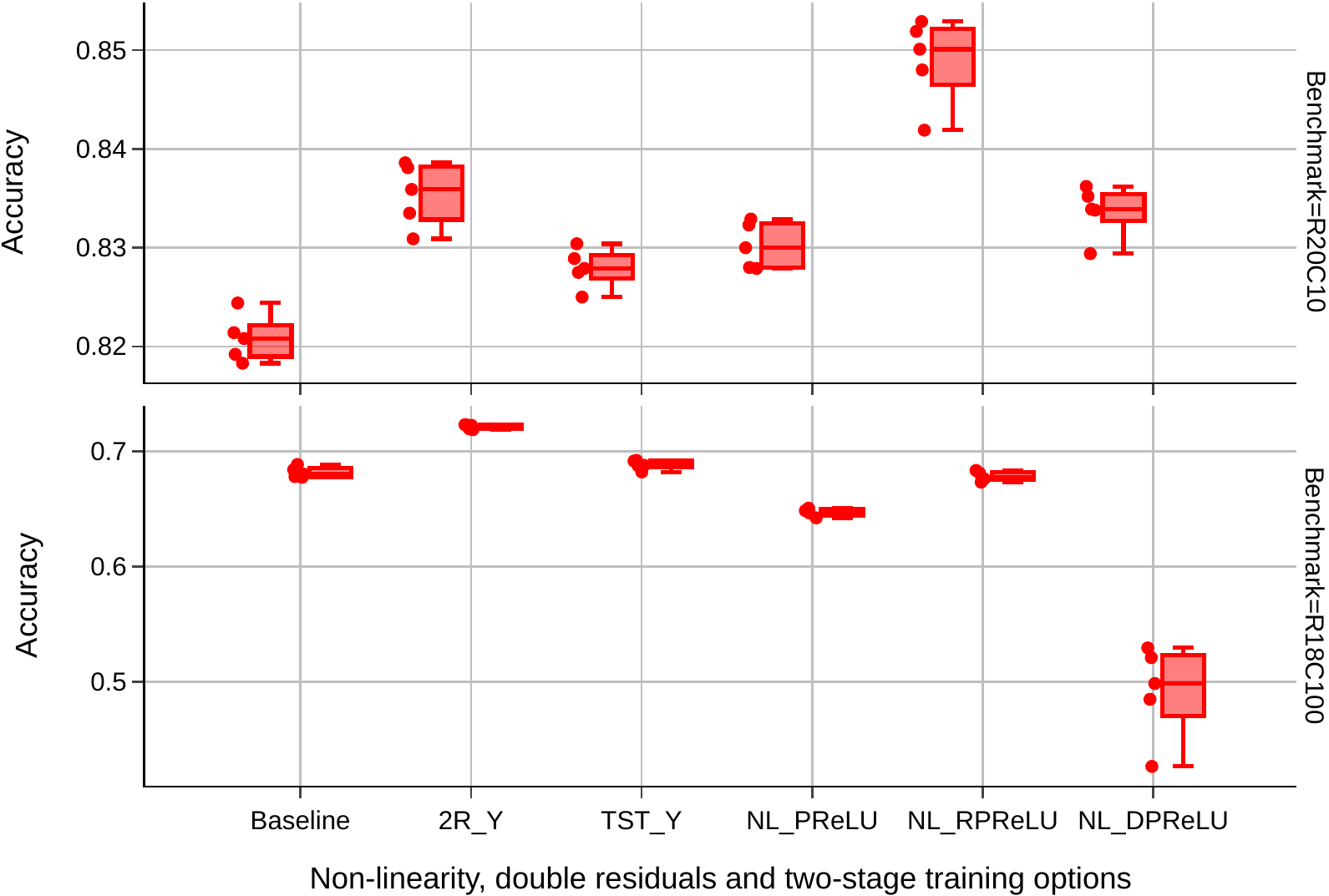}
  \caption{Activation function (non-linearity), double residual and two-stage training repair methods compared to the baseline.}
  \label{fig:dse_other}
\end{figure}
Fig. \ref{fig:dse_other} shows the remaining accuracy repair categories: activation function, double residual, and two-stage training. The activation function category shows different results in the R20C10 and R18C100 benchmarks. For R20C10 an additional non-linearity helps a lot, up to 3\%-point, whereas R18C100 shows no improvement or even a reduction in accuracy. 
In both benchmarks, we see that RPReLU results in higher accuracy than the regular PReLU. This is reasonable since the RPReLU adds additional learning capacity to the network. Additionally, in both benchmarks, we see that DPReLU performs worse than RPReLU even though DPReLU is a superset of RPReLU. 

Employing two residual connections instead of one results in an improvement in both benchmarks, 1.4\%-point in R20C10 and 3.5\%-point in R18C100. Since this repair method reduces the lifetime of higher-precision values, it could be very suitable for BNNs on embedded systems.

The two-stage training scheme has a negligible increase in accuracy, at most 0.5\%-point. What could be beneficial in this method is to change the division in training time between the first and second stage. In our experiments, the first stage with real-valued weights and binary activations more quickly converges and therefore this method could be used to speed-up training when the amount of training time of the first stage is reduced.

\section{Discussion and future research} \label{sec:discussion}
BNNs save on memory and simplify compute logic by binarizing both features and weights. This extreme form of quantization inevitably causes accuracy loss. To reduce the accuracy gap many repair techniques have been proposed. As outlined in the overview, BNN papers do not agree on what series of repair methods are best to use. Therefore, in the previous section, we have conducted a design space exploration of each accuracy repair technique in isolation and compared it against a baseline BNN. Section \ref{sec:discussion:accgap} discusses this study and the remaining accuracy gap, while Section \ref{sec:discussion:cost} raises a discussion about the benefit and cost of BNN network architectures.

\subsection{Accuracy gap} \label{sec:discussion:accgap}
From the study in Section \ref{sec:review}, we found that some repair categories and methods are more beneficial than others. The most important categories are feature binarizer, feature normalization, and double residual. 

Within the feature binarizer category, the LC (with a larger clipping interval than 1, but smaller than 3) and PN binarizers performed best with up to a 3\%-point increase. An interesting observation is that these fixed-form binarizers outperformed binarizers that gradually changed shape during training. Besides that, since the best binarizers differ in shape and clipping interval, further research is needed to find a binarizer that combines the two.

Whereas the clipping interval can be fixed, the best feature normalization method implicitly changed the interval using a division by the standard deviation (STD). This resulted in an increase in the baseline's accuracy in both benchmarks of up to 2\%-point. Since this method has a good improvement and implicitly affects the binarizer, we feel that future research should look at the interdependence between binarizers and normalization.

While the two-stage training repair method did not result in any major accuracy improvement, we found that this method could potentially be used in the future to speed-up BNN training. Research should focus on the amount of training time between the first and second stage. The first stage should be used to quickly find an appropriate neighborhood of minima, while the second stage takes care of actual convergence to a minimum when both features and weights are binarized.

Besides training techniques, there is one network topology repair method that results in a large improvement as well: the double residual. This method resulted in an improvement of up to 3.5\%-point. 

Overall, any individual repair method does not reach the highest accuracy as reported in the literature. Roughly there is still a 3\%-point gap in both benchmarks between the best accuracy reported BNN in the literature and our best results. Since we only evaluated individual repair methods and BNNs from literature employ multiple repair methods, it implies that the improvements by repair methods are most likely partly additive. Thus, future research should focus on combining successful methods to see what combinations are most beneficial and why.

To a great degree, the accuracy difference between binary and full-precision networks can be reduced by employing multiple repair methods, as can be seen in our literature overview. However, a gap of approximately 3\%-point remains to be solved. We believe that future research closes the gap by introducing new training techniques and/or by finding more appropriate network topologies.

\subsection{Benefit and cost of BNNs} \label{sec:discussion:cost}
While the benefits of repair methods are obvious, accuracy increases, we did not look at the cost of certain methods when performing inference on an embedded system. A cost can be defined as additional energy consumption through more memory or compute usage, and/or as additional hardware area required. In our classification of accuracy repair methods, there were two main branches of repair techniques: training techniques and network topology changes. The former branch does not introduce any inference cost, whereas any method in the latter branch does. In general, any method in this branch adds high-precision operations and potentially high-precision data movement between memory and processing elements, which will increase energy consumption. Moreover, the hardware needs to support mixed-precision operations, which usually implies additional area is required for having multiple units with varying precision. Note that the ResNet-architecture, on which most BNNs are based, requires mixed-precision support due to the high-precision residual connections. The effect on energy consumption of high-precision operations and residuals in a BNN is illustrated and quantified by \cite{Guo2021BoolNet:Networks}. They claim an energy reduction of a factor 6 by the removal of all high-precision operations at the cost of a 17\%-point accuracy drop.

One might wonder, whether binary is the way to go. One alternative to binary could be ternary quantization which adds a 0 to the quantization possibilities: $[-1,0,+1]$. This extra form of precision results in higher accuracy while coming with extra hardware benefits like zero-suppression. As illustrated by \cite{Scherer2021CUTIE:Efficiency}, an energy efficiency can be achieved that is better than binary. Looking at high-performance CNN accelerators, like the TPU \cite{Jouppi2017In-datacenterUnit}, Google has switched from 8-bit integers in its first version to 16-bit brainfloats in its second version. This format is quite convenient since it allows conversion of 32-bit floating-point to 16-bit brainfloat, thereby enabling model portability between devices and simplifying the optimization process as both training and inference are run with similar precision.

\section{Conclusion} \label{sec:conclusion}
Many embedded systems became intelligent by applying some form of deep learning, often using CNNs. However, the deployment of state-of-the-art CNNs on resource-constrained embedded systems is not straightforward. CNNs are computationally and memory intensive because of their billions of parameters and operations. To enable the deployment of CNNs heavy optimization is crucial. A key optimization is model compression by quantization. Using the extreme form of quantization, binarization, features, and weights of a CNN are compressed to a single bit, thereby heavily saving upon memory size and throughput, and simplifying compute logic. In other words, BNNs are the key enabler of CNNs on embedded systems. Unfortunately, BNNs suffer from severe accuracy degradation. To reduce the accuracy gap between binary and full-precision networks, many repair methods have been proposed in the recent past, which we have classified and put into a single overview in this chapter. The repair methods are divided into two main branches, training techniques and network topology changes, which can further be split into smaller categories. The training techniques categories are binarizer implementation, normalization, teacher-student, regularization, two-stage training, and optimizer, while the network topology categories are: scaling factor, non-linearity, ensemble, double residual, and squeeze-and-excitation.

In the overview, we see that progress has been made in reducing the accuracy gap, but BNN papers are not aligned on what repair methods should be used to get highly accurate BNNs. Therefore, we performed an empirical review that evaluates the benefits of each repair method in isolation over the ResNet-20\&CIFAR10 and ResNet-18\&CIFAR100 benchmarks. We found three repair categories most beneficial: feature binarizer, feature normalization, and double residual. Within the feature binarizer category, the LC (with larger clipping interval) and PN binarizers performed best with up to a 3\%-point increase. Regarding feature normalization division by standard deviation worked best resulting in at most a 2\%-point increase. Next to these training techniques, the double residual that changes the network topology also gave a performance boost of at most 3\%-point.

While these individual repair methods achieve an accuracy increase, there is still an accuracy gap between the binary and full-precision network and therefore more research is needed. Future research could focus on the interdependence of repair methods and/or new network topologies. Additionally, future research should take the cost (energy or area) of a repair method into account.

In reality, it remains to be seen whether BNNs will be able to close the accuracy gap while staying highly energy-efficient on resource-constrained embedded systems.

\newpage
\bibliographystyle{spmpsci_natbib.bst}
\bibliography{references}

\end{document}